 \documentclass[pmlr,twocolumn,10pt]{jmlr} 

\usepackage{booktabs}
\usepackage{siunitx}
\usepackage{enumitem}
\usepackage{makecell}
\usepackage{multirow}

\usepackage[switch]{lineno}

\theorembodyfont{\upshape}
\theoremheaderfont{\scshape}
\theorempostheader{:}
\theoremsep{\newline}

\jmlrvolume{259}
\jmlryear{2024}
\jmlrsubmitted{LEAVE UNSET}
\jmlrpublished{LEAVE UNSET}
\jmlrworkshop{Machine Learning for Health (ML4H) 2024} 

 \title[A Study on Context Length and Efficient Transformers for Biomedical Image Analysis]{A Study on Context Length and Efficient \\ Transformers for Biomedical Image Analysis}

\author{%
\Name{Sarah Hooper} \\
\addr Office of AI Research, NHLBI, NIH, Bethesda, MD, USA
\AND
\Name{Hui Xue} \\
\addr Health Futures, Microsoft Research, Redmond, WA, USA
}

\begin{document}

\maketitle

\begin{abstract}

Biomedical images are often high-resolution and multi-dimensional, presenting computational challenges for deep neural networks. These computational challenges are compounded when training transformers due to the self-attention operator, which scales quadratically with context length. 
Recent works have proposed alternatives to self-attention that scale more favorably with context length, alleviating these computational difficulties and potentially enabling more efficient application of transformers to large biomedical images. However, a systematic evaluation on this topic is lacking.
In this study, we investigate the impact of context length on biomedical image analysis and we evaluate the performance of recently proposed substitutes for self-attention.
We first curate a suite of biomedical imaging datasets, including 2D and 3D data for segmentation, denoising, and classification tasks. We then analyze the impact of context length on network performance using the Vision Transformer and Swin Transformer.
Our findings reveal a strong relationship between context length and performance, particularly for pixel-level prediction tasks. 
Finally, we show that recent attention-free models demonstrate significant improvements in efficiency while maintaining comparable performance to self-attention-based models.  

\end{abstract}

\paragraph*{Data and Code Availability}
Code will be available on GitHub. Five of the datasets are public datasets; the cardiac MR denoising dataset is a private dataset that is not currently available externally.

\paragraph*{Institutional Review Board (IRB)}
This study did not require IRB approval.

\section{Introduction}
\label{sec:intro}

Biomedical and clinical imaging modalities often produce high-resolution, multi-dimensional images that contain rich and detailed information. These large image sizes present computational challenges for deep neural networks, such as increased memory requirements and long processing times \citep{dinsdale2022challenges, suzuki2017overview, berisha2021digital}.

The popularity of transformers has compounded the computational difficulties of training neural networks on medical images. Central to transformers is the self-attention operator, which scales quadratically with context length \citep{keles2023computational}. This quadratic scaling can be prohibitive when training models on medical images, where capturing fine-grained details in high-resolution, multi-dimensional images is critical. 

\begin{figure*}[!t]
\floatconts
  {fig:context_length}
  {\caption{Visualization of how context length changes with patch size and attention window size. 
  When using ViT, we use smaller patches to tokenize the input image, resulting in longer context lengths. When using Swin, we use larger windows of attention, resulting in longer context lengths.
  }}
  {\includegraphics [width=.9\textwidth]{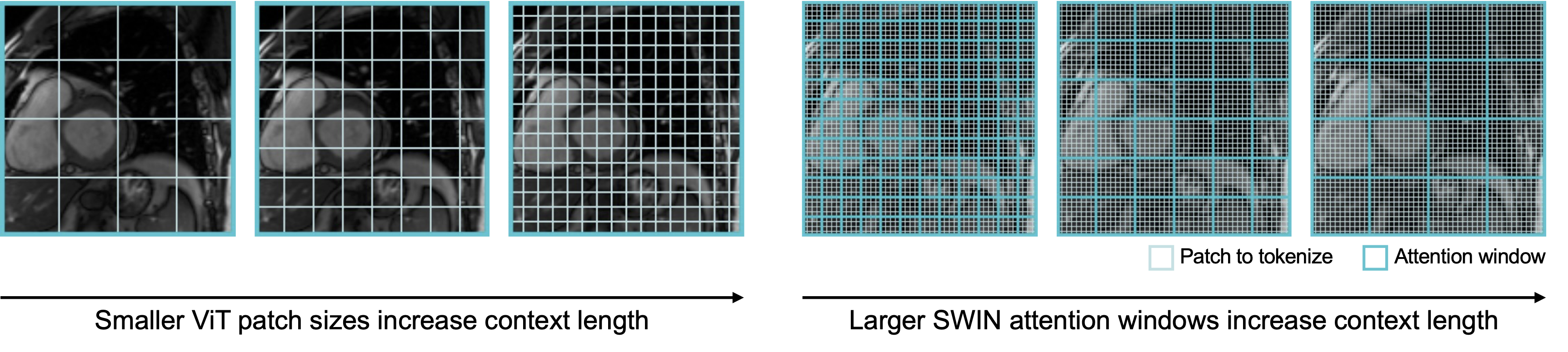}}
\end{figure*}

In natural language processing (NLP), recent efforts have improved the efficiency of self-attention \citep{dao2022flashattention, beltagy2020longformer, child2019generating, katharopoulos2020transformers, choromanski2020rethinking,DBLP:journals/corr/abs-2009-06732} or have investigated replacing it all together \citep{gu2021efficiently, poli2023hyena, peng2023rwkv, fu2022hungry, sun2023retentive, gu2023mamba}. These works aim to design operators that match the performance of self-attention while scaling more favorably with context length, enabling models to process longer inputs. Such advances have gained popularity in NLP, driving new innovation and capabilities \citep{dong2023survey, tsirmpas2024neural, huang2023advancing, pawar2024and}. While such long-context models also hold promise for biomedical image analysis---potentially making transformers more efficient and effective when applied to high-resolution images---a systematic study on this topic is lacking.

In this work, we investigate long-context models for biomedical imaging. We ask two questions: do medical imaging applications benefit from longer context, and if so, what are efficient and effective approaches for training long-context models? We present a thorough investigation on the impact of context length on imaging applications and assess the performance of recently proposed alternatives to self-attention. 

We begin by curating a suite of biomedical imaging datasets comprising both two- and three-dimensional data as well as common medical imaging tasks: segmentation, image denoising, and classification. 
Including these diverse data and task types enables us to evaluate long-context models in different settings.

We then examine how varying context length impacts performance on these tasks using common transformers for computer vision. We evaluate the impact of patch size on the vision transformer (ViT, \cite{dosovitskiy2020image}) and the impact of the attention window on the Swin transformer \citep{liu2021swin}---both of which increase transformer context length (Figure \ref{fig:context_length}). 
We find a strong relationship between patch size and performance, particularly for pixel-level prediction tasks (e.g., denoising).

Finally, we evaluate recently proposed alternatives to self-attention (Hyena \citep{poli2023hyena} and Mamba \citep{gu2023mamba}) to evaluate how each impacts performance and efficiency. Our results show these operators can achieve comparable performance to self-attention while improving efficiency by over 80\%, underlining the importance of efficient long-context processing for biomedical imaging. 

\section{Related Work}
\label{sec:rw}

\paragraph{Vision Transformers.} The transformer, initially introduced for NLP \citep{vaswani2017attention}, has been widely adapted and applied to vision tasks. ViT showed that a transformer architecture nearly identical to those used in NLP achieved strong performance on image recognition \citep{dosovitskiy2020image}. Follow-on works adapted the transformer for specific vision tasks \citep{han2022survey, khan2022transformers, shamshad2023transformers}. For example, Swin introduced a shift-and-merge windowing scheme, wherein image patches only attended to local windows, reducing computational complexity and improving performance on pixel-level prediction \citep{liu2021swin}. Similarly, PVT and Segformer introduced hierarchical transformer architectures designed for dense prediction tasks \citep{wang2021pyramid, xie2021segformer}. Finally, work like DeiT introduced training and distillation strategies to improve the data efficiency of vision transformers \citep{touvron2021training}. 

\paragraph{Efficient Attention.} While transformers achieve strong performance, their self-attention operator scales quadratically with context length \citep{keles2023computational}, leading to prohibitive computational demands for processing long-context inputs. In response, many works have proposed approaches to improve attention's efficiency. 
Flash attention is a popular approach that is an exact, hardware-aware implementation of attention, reproducing attention but with subquadratic scaling \citep{dao2022flashattention, dao2023flashattention, shah2024flashattention}. 
Other approaches propose approximations to attention, including sparse and local attention \citep{beltagy2020longformer, child2019generating}, linear attention \citep{katharopoulos2020transformers}, and others \citep{choromanski2020rethinking,DBLP:journals/corr/abs-2009-06732}. These approaches are more efficient than self-attention, but typically trade-off speed with expressivity and performance \citep{poli2023hyena}.

\paragraph{Alternatives to Attention.} An alternative approach to making attention more efficient is to replace it entirely \citep{poli2023hyena, peng2023rwkv, fu2022hungry, nguyen2022s4nd, sun2023retentive}. This class of approaches tries to construct operators that maintain attention's performance while scaling more favorably with context length. For example, the Hyena operator leverages long convolutions to match self-attention's ability to capture global dependencies but with an operation that scales subquadratically with context length \citep{poli2023hyena}. Other approaches include state space models (SSMs), which take inspiration from traditional signal processing models \citep{gu2021efficiently, gu2021combining}. \cite{gu2023mamba} recently proposed the selective SSM in a model called Mamba, which increases the expressivity of SSMs and achieves promising performance on NLP and audio tasks.

Some of these alternatives have been evaluated for vision tasks. For example, early SSM models were adapted to image classification \citep{nguyen2022s4nd}, Hyena showed proof-of-principal on ImageNet \citep{poli2023hyena}, and Mamba has been adapted for natural image processing \citep{zhu2024vision, liu2024vmambavisualstatespace}. Similarly, related work has proposed new architectures leveraging some of these efficient operators for medical applications \citep{fillioux2023structured,archit2024vim, xing2024segmamba, wang2024mamba, ma2024u,nasiri2024vim4path},
however these applications typically focus on a single task and architecture instead of a systematic evaluation over many operators, tasks, and data types.

\paragraph{Image Resolution and Context Length.}
There is a growing body of evidence that context length and image resolution play key roles in the quality of representations learned by transformers. While not synonymous, image resolution and context length are closely linked, as smaller patches used to tokenize the image better preserve image resolution at the expense of increased context length (Figure \ref{fig:context_length}).

For example, a study on masked autoencoding showed improved performance for increasing context length \citep{hu2022exploring}.
Diffusion models have shown improved performance with decreased patch size \citep{peebles2023scalable}. 
A recent work showed competitive performance tokenizing images at the pixel-level \citep{nguyen2024image}, a finding consistent with the results of this work and which further motivates our exploration of efficient alternatives to attention. 
Recent work in multimodal pretraining have found improved performance with higher-resolution images \citep{meng2024deepstack, mckinzie2024mm1}. 
A few studies have looked at the impact of ViT patch size on classification, finding improved performance with smaller patches \citep{than2021preliminary, ibrahimovic2023optimizing, beyer2023flexivit}. 
Finally, prior work has explored conceptually similar questions using CNNs. For example, several studies have highlighted the importance of preserving image resolution to achieve high CNN performance \citep{thambawita2021impact, sabottke2020effect}, and some work has suggested larger convolutional filter sizes improve CNN performance \citep{ding2022scaling}.

\paragraph{Summary.}
While significant progress has been made improving transformer efficiency for long-context inputs in NLP, a systematic evaluation of the relationship between context length, efficiency, and performance in biomedical imaging is lacking. Further, many efficient operators have not been tested in common medical imaging settings (e.g., with 3D data, for improving image quality).
We aim to fill these gaps by investigating the impact of context length and the performance of efficient attention alternatives on diverse biomedical imaging datasets, offering insights into the development of more efficient deep learning models for biomedical applications.

\section{Approach}
\label{sec:approach}

We begin with background on self-attention and the alternative operators we evaluate. We then discuss model architectures, our approach to changing context length, and our evaluation datasets.

\begin{figure*}[htbp]
\floatconts
  {fig:operators}
  {\caption{Attention and alternative operators. Left, we show a standard transformer block. Right, we show the operators we evaluate in the transformer blocks: self-attention, Hyena, and MambaVision.}}
  {\includegraphics [width=.9\textwidth]{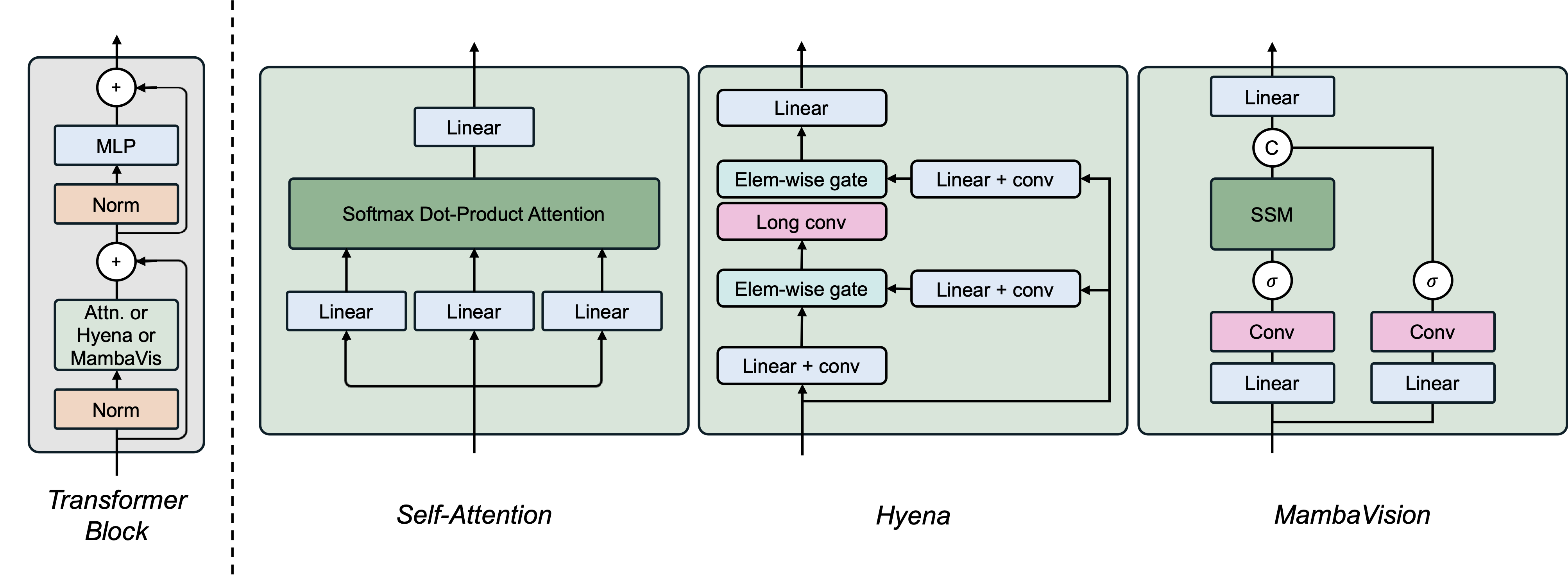}}
\end{figure*}

\subsection{Background: Attention and Alternatives}

\paragraph{Self-Attention}

We show the standard transformer block in Figure \ref{fig:operators}, which is traditionally powered by self-attention \citep{vaswani2017attention, dosovitskiy2020image}. 
For an input sequence $X \in \mathbb{R}^{n \times d}$, 
where $n$ is the sequence length and $d$ is the sequence dimension, 
self-attention maps this sequence to $Y \in \mathbb{R}^{n \times d}$ 
using the set of trainable parameters 
$W_q \in \mathbb{R}^{d \times d}$, 
$W_k \in \mathbb{R}^{d \times d}$, 
$W_v \in \mathbb{R}^{d \times d}$. 
First, the query, key, and value matrices are computed as 
$Q = XW_q$, $K = XW_k$, and $V = XW_v$. 
The softmax dot-product self-attention operation is then defined as:

\begin{gather*}
\text{Attention}(Q, K, V) = \text{Softmax} \left( \frac{QK^\top}{\sqrt{d}} \right) V.
\end{gather*}

The computational complexity of self-attention is $O(n^2)$ \citep{keles2023computational}, meaning using self-attention with longer sequences results in quadratic increases to memory and computation. 

\paragraph{Alternatives to Attention.}
Many alternative operators have been proposed to enable longer context processing. To do a thorough analysis across tasks, datasets, and context lengths, we carefully selected which alternatives to evaluate. We selected operators that showed proof-of-principal performance on imaging tasks and outperformed similar baselines. Further, we selected operators that could be swapped out for attention in existing architectures, enabling a direct comparison between operators without confounding influences from other architectural changes.

\paragraph{Hyena.} We selected the Hyena operator as the first attention alternative to evaluate \citep{poli2023hyena} (Figure \ref{fig:operators}). Hyena uses long convolutions to achieve subquadratic scaling with respect to context length, while still maintaining token-level precision and global context. Hyena further introduces element-wise gating to inject data dependence into the operator, mimicking the data dependence property of self-attention. 
The computational complexity of Hyena is $O(nlog_2(n))$ \citep{poli2023hyena}.

We selected Hyena because it maintains two characteristics of attention---token-level precision and global context---that we hypothesized would help maintain performance on both sparse and dense image analysis tasks. Additionally, Hyena has shown strong performance on ImageNet and has exceeded the performance of or generalized related methods \citep{nguyen2022s4nd,fu2022hungry,poli2023hyena}.

\paragraph{Mamba.} We selected MambaVision as the second operator to evaluate. Mamba is a selective SSM that transforms an input $X$ into output $Y$ via a learnable hidden state \citep{gu2023mamba}. We evaluated the MambaVision operator proposed by \cite{hatamizadeh2024mambavision}, which adapts the selective SSM module in \cite{gu2023mamba} to vision tasks. MambaVision incorporates a selective SSM along with a skip connection (Figure \ref{fig:operators}), defined as:

\begin{gather*}
Z_1 = Scan(\sigma(Conv(Linear_{d \rightarrow \frac{d}{2}}(X)))) \\
Z_2 = \sigma(Conv(Linear_{d \rightarrow \frac{d}{2}}(X))) \\
Y = Linear_{\frac{d}{2} \rightarrow d}(Concat(Z_1, Z_2))
\end{gather*}

where $Scan(\cdot)$ is the selective scan operation in \cite{gu2023mamba} and $\sigma$ is the SiLU function.

We selected Mamba as a SotA SSM approach that has been adapted to vision with promising initial results. 
Further, MambaVision reportedly exceeds the performance of other Mamba vision architectures \citep{liu2024vmambavisualstatespace, zhu2024vision, pei2024efficientvmamba}.

\begin{figure*}[!t] 
\floatconts
  {fig:tasks}
  {\caption{Task visualization. We visualize a network input and ground truth output for each task. Starting from the upper left and moving clockwise: retinal vessel segmentation, microscopy denoising, pneumothorax classification, pulmonary embolism classification, CMR denoising, and abdominal CT organ segmentation.}}
  {\includegraphics [width=.9\textwidth]{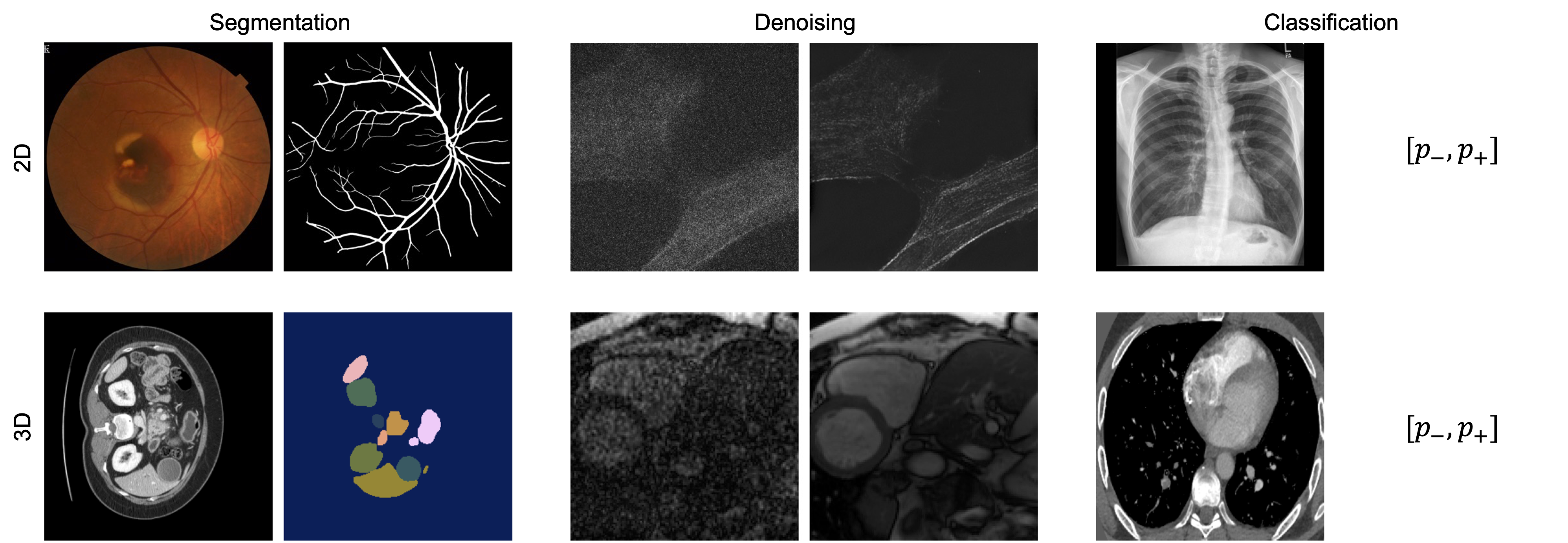}}
\end{figure*}

\subsection{Model Architectures}

We evaluated two widely used architectures for vision: ViT \citep{dosovitskiy2020image} and Swin \citep{liu2021swin}. ViT closely mirrors transformers used in NLP. Swin restricts attention to local windows, then shifts and merges these windows. By stacking multiple Swin transformer blocks, the effective receptive field grows.
 To keep the number of parameters similar between ViT and Swin, we used ViT's \textit{small} configuration and Swin's \textit{tiny} configuration. 
 
 We selected ViT and Swin as two common vision transformers used in medical imaging applications \citep{he2023transformers, shamshad2023transformers} that other transformers share similarities with. For example, DeiT’s architecture is nearly identical to ViT, while PVT and Segformer compress patches in attention-based blocks, similar to Swin. 

Both ViT and Swin are made up of repeating transformer blocks. Traditionally, these blocks are powered by self-attention. We evaluated attention as well as Hyena and MambaVision when used as drop-in replacements for attention, as shown in Figure \ref{fig:operators}.\footnote{We removed Swin's shift operation when using Hyena and MambaVision, as the masking procedure used with attention does not translate to the alternative operators. 
We evaluate the impact of the shift operator in the Appendix.}

For classification tasks, we used a linear layer as the task head. For pixel-level prediction tasks, we used the ViT UNETR head \citep{hatamizadeh2022unetr} for ViT and the UPerNet head \citep{xiao2018unified} for Swin. We chose these prediction heads as they are relatively lightweight and maintain similar parameter counts between ViT and Swin models. 

\subsection{Changing Context Length}

Consistent with most transformers for computer vision, both ViT and Swin begin with a patch embedding layer that partitions the image into non-overlapping patches, which are then embedded and used as tokens. The context length of the self-attention operator is defined by how many tokens are processed concurrently. Thus, longer context lengths occur when attending to more image patches. 

We can vary context length by (i) changing the patch size, thereby increasing the number of tokens per image region; or (ii) changing the size of the attention window, enabling attention among a greater portion of the image. We explore both in this work.

To change the context length in ViT, we swept the patch size used in the patch embedding layer. We evaluated 32-, 16-, 8-, and 4-pixel isotropic patches. Reducing the patch size increases context length and computational complexity, but results in a higher resolution representation of the input image (Figure \ref{fig:context_length}). 

For Swin, we fixed the embedding patch size to 2-pixel isotropic patches while we varied the size of the local attention window. We evaluated 4-, 8-, and 16-token isotropic windows. Larger windows increase context length and computational complexity, but enable the network to use a greater portion of the image to inform each token's representation (Figure \ref{fig:context_length}). In the Appendix, we also evaluate the impact of the patch size on Swin performance.
 
These changes to context length do not strongly impact the parameterization of the attention modules. However, changing ViT's patch size does change the number of parameters in the patch embedding layer. We provide parameter counts in the Appendix.

\begin{figure*}[!t]
\floatconts
  {fig:vit_performance}
  {\caption{ViT performance. We visualize performance for each task, operator, and patch size with 95\% confidence intervals. An X on the x-axis indicates the patch size exceeded our hardware capacity.}}
  {\includegraphics [width=.9\textwidth]{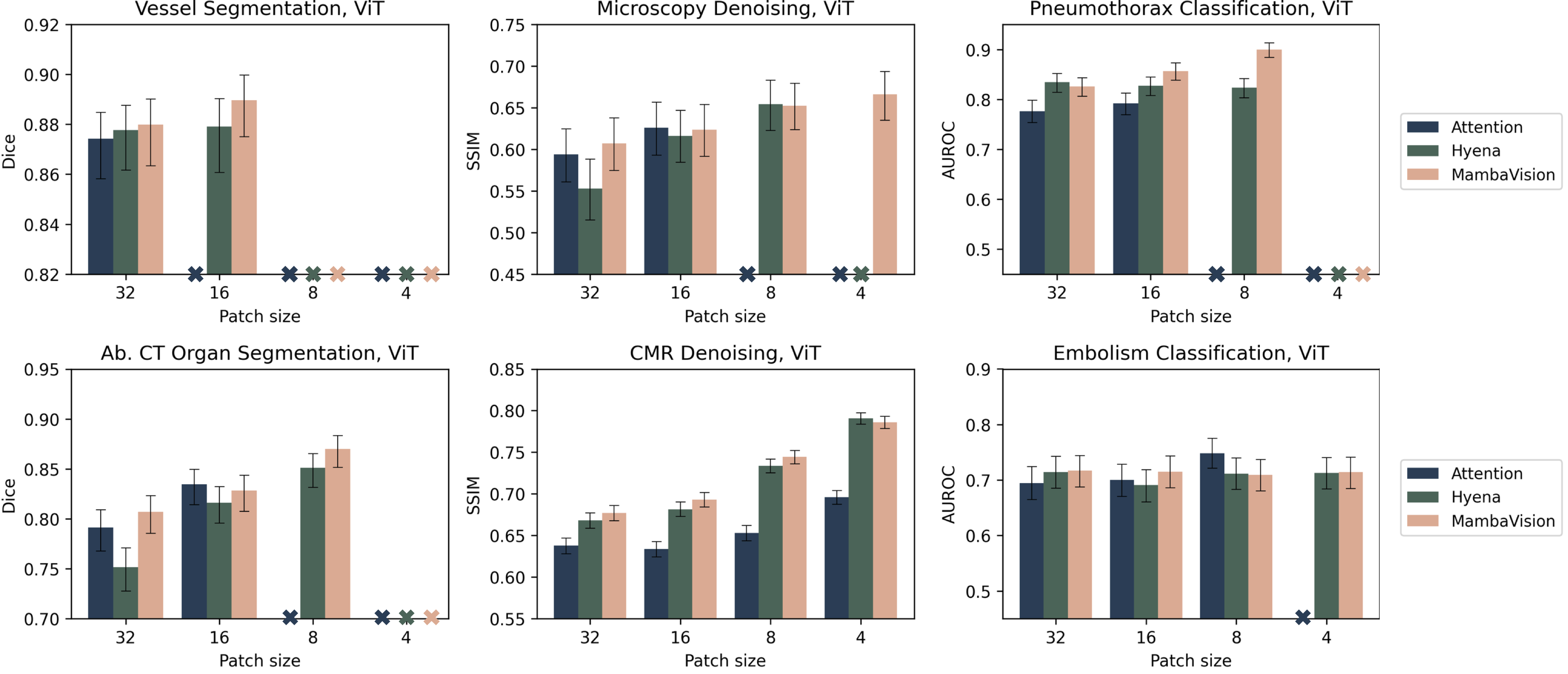}}
\end{figure*}

\subsection{Dataset and Task Selection}

We selected diverse biomedical imaging tasks to evaluate the impact of context length and self-attention. We included segmentation to evaluate the networks' ability to identify pixel-level features. We included image denoising as a task that requires models to restore high-fidelity details. Finally, we included classification to evaluate the networks' ability to aggregate global information and predict image-level labels. 
For each task type, we included 2D and 3D data from different imaging modalities.
This comprehensive evaluation allowed us to analyze how context length and different operators influence performance across many datasets as well as tasks that require fine-grained precision and global understanding. 

Our tasks are visualized in Figure \ref{fig:tasks} and described below, with additional details in the Appendix.
\begin{itemize}[leftmargin=*, itemsep=0pt]
        \item 2D Retinal Vessel Segmentation. This public fundus photograph dataset contains 800 images, each of shape $2048 \times 2048$ pixels with three channels \citep{jin2022fives}. Each image has pixel-wise annotations of retinal vessels.
        \item 3D Abdominal CT Organ Segmentation. This public dataset contains 945 images, each with nine organs segmented \citep{qu2024abdomenatlas, antonelli2022medical}. We resized each axial slice to $256 \times 256$ pixels and cropped to 64 axial slices per volume. 
        \item 2D Microscopy Denoising. This public fluorescence microscopy dataset contains 360 images, each of shape $1024 \times 1024$ \citep{zhou2020w2s}. Each sample contains a paired high- and low-SNR image. 
        \item 3D Cardiac MRI (CMR) Denoising. This private dataset contains 13,964 retro-gated cines, each with 32 frames and center cropped to $128 \times 128$ pixels. Each sample contains a paired high- and low-SNR image.
        \item 2D Pneumothorax Classification. This public chest x-ray dataset contains 18,887 chest x-rays, each of $1024 \times 1024$ pixels \citep{feng2021curation}. 15\% of the images contain a pneumothorax. 
        \item 3D Pulmonary Embolism Classification. This public CT dataset contains 7,205 images, 32\% positive for pulmonary embolism \citep{colak2021rsna}. We resized each axial slice to $256 \times 256$ pixels and cropped to 64 axial slices per volume. 
\end{itemize}

\section{Experiments}
\label{sec:experiments}

We first describe our experimental setup, then evaluate task performance and training efficiency as a function of context length.

\subsection{Experimental Setup}

We split the datasets randomly by patient into 60\% train, 20\% validation, and 20\% test, except for the vessels dataset which had pre-defined splits. We tuned the learning rate for each experiment; final learning rates are given in the Appendix.

We trained the classification and segmentation tasks using the cross entropy loss and the denoising tasks using the sum of the mean squared error loss, Charbonnier loss, and Gaussian loss. 
We used an affine transform and brightness jitter as training augmentations for all tasks except CMR denoising, where we only used an affine transform. We did not use brightness jitter on CMR denoising since the pixel values are representative of the SNR. 

Other training parameters were kept constant for all experiments. We used the Adam optimizer with a one cycle learning rate scheduler and no weight decay. All experiments were run for 250 epochs on eight 80GB NVIDIA A100s using Python 3.11. Models were checkpointed using the minimum validation loss.

\begin{figure*}[htbp]
\floatconts
  {fig:swin_performance}
  {\caption{Swin performance. We visualize performance for each task, operator, and patch size with 95\% confidence intervals. An X on the x-axis indicates that the window size exceeded our hardware capacity.}}
  {\includegraphics [width=.9\textwidth]{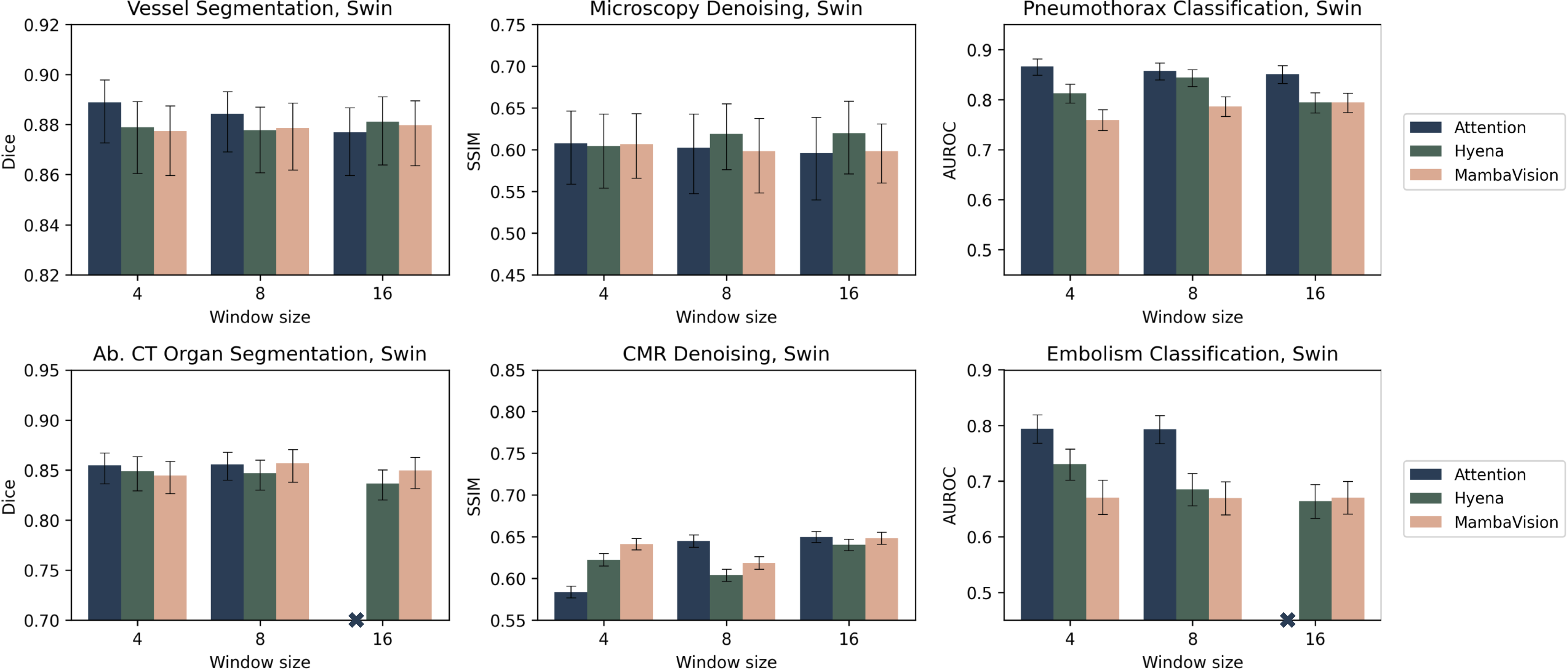}}
\end{figure*}

\subsection{Task Performance}
We next report the task performance for each network with changing context lengths and operators, as shown in Figures \ref{fig:vit_performance} and \ref{fig:swin_performance}. We evaluated segmentation performance using the Dice coefficient, denoising performance using the structural similarity index measure (SSIM), and classification performance using the area under the receiver operating curve (AUROC). We computed 95\% confidence intervals by bootstrapping over the test set. 

\paragraph{Patch Size Strongly Impacts ViT Performance.}
In Figure \ref{fig:vit_performance}, we observe a strong relationship between patch size and performance. Using self-attention, the best performance across all tasks was achieved by the smallest patch size.

We notice a particularly strong correlation for pixel-level prediction, with all operators consistently achieving improved performance across the four pixel-level prediction tasks with smaller patch sizes. 

The trend on classification is less pronounced, although attention-based networks still saw improved performance with decreasing patch size, with an average 4.85\% increase in performance comparing the largest and smallest patch size. 

In the Appendix, we further evaluate the impact of patch size on Swin performance to verify we observe the same trends shown above with ViT. To summarize our findings, we observed an average 8.66\% improvement to performance using 2-pixel isotropic patches instead of 4-pixel isotropic patches in Swin, with performance improving across all of our six tasks with the smaller patch size. 
These results indicate that preserving resolution via smaller patch sizes is important to performance in both architectures. In the remainder of the main text, we evaluate Swin with 2-pixel isotropic patches.

\paragraph{Attention Window Size has Only Minor Impacts on Swin Performance.} We do not observe a strong relationship between the attention window size and Swin performance (Figure \ref{fig:swin_performance}). 
While CMR denoising performance improved with larger windows in attention-based networks---with 16-token windows improving performance 11.37\% compared to 4-token windows---we observed only minor differences for segmentation and classification, with performance sometimes decreasing. 
The improved performance in the CMR denoising task might be attributed to the dataset containing videos, as increasing the window size provides the network with additional frames of the same structure to aid in the denoising process.
For other tasks, local information captured in small windows combined with Swin's window merging may provide a sufficient balance of local and global information to achieve high performance.

\paragraph{Attention Alternatives Achieve Strong Performance.} Both Hyena and Mamba showed promising performance. We summarize their change in performance compared to attention given the same network configuration in Table \ref{tab:perf_change_vit} and Table \ref{tab:perf_change_swin}. 

\begin{table}[hbtp]
\floatconts
  {tab:perf_change_vit}
  {\caption{Average ViT performance change compared using self-attention.}}
    {
    \centering
    \small
    \begin{tabular}{c|ccc}
    \toprule
     \makecell{Performance \\ change, ViT}
     & Segment
     & Denoise
     & Classify \\
    \midrule
    Hyena & 
    -2.26\% & 4.96\% & 1.69\% \\
    MambaVision & 
    0.64\% & 7.38\% & 2.91\% \\
    \bottomrule
    \end{tabular}}
\end{table}

\begin{table}[hbtp]
\floatconts
  {tab:perf_change_swin}
  {\caption{Average Swin performance change compared to using self-attention.}}
    {
    \centering
    \small
    \begin{tabular}{c|ccc}
    \toprule
     \makecell{Performance \\ change, Swin}
     & Segment
     & Denoise
     & Classify \\
    \midrule
    Hyena & 
    -0.61\% & 0.85\% & -7.19\% \\
    MambaVision & 
    -0.54\% & 0.85\% & -11.68\% \\
    \bottomrule
    \end{tabular}}
\end{table}

While there is a consistent performance gap between attention and the alternatives on Swin classification, part of the differential is likely due to the absence of the shift operation with the alternative operators (see Appendix for more details).

\subsection{Training Efficiency}
We next evaluate training efficiency. 
While smaller patches can improve performance, they also increase computational complexity due to increased context length. For example, when training a self-attention-based ViT on our datasets, using 16- or 8-pixel patches increased the time required for a forward and backward pass by 252.90\% and 2,335.48\% compared to using 32-pixel patches, respectively. This drastic increase in computation with longer context lengths motivates the use of more efficient operators. 

To assess the efficiency of each model, we evaluated the time required to perform a forward and backward pass as well as the maximum memory allocated. 
We provide results for all runs in the Appendix and summarize key findings in Tables \ref{tab:vit_speedup} and \ref{tab:swin_speedup}, where we report the average speedup achieved by Hyena and MambaVision compared to attention.

\paragraph{Attention Alternatives Improve Efficiency at Long Context Lengths.} 

\begin{table}[hbtp]
\floatconts
  {tab:vit_speedup}
  {\caption{Average ViT speedup compared to networks that use self-attention.}}
    {
    \centering
    \small
    \begin{tabular}{c|cccc}
    \toprule
     \shortstack{Speedup, ViT}
     & \shortstack{Patch\\32}
     & \shortstack{Patch\\16}
     & \shortstack{Patch\\8}
     & \shortstack{Patch\\4} \\
    \midrule
    Hyena & 
    -48.66\% & 5.50\% & 42.79\% & 81.49\% \\
    MambaVision & 
    -7.68\% & 32.67\% & 57.74\% & 86.82\% \\
    \bottomrule
    \end{tabular}}
\end{table}
 
\begin{table}[hbtp]
\floatconts
  {tab:swin_speedup}
  {\caption{Average Swin speedup compared to using self-attention.}}
    {
    \centering
    \small
    \begin{tabular}{c|cccc}
    \toprule
     \shortstack{Speedup, Swin}
     & \shortstack{Window\\4}
     & \shortstack{Window\\8}
     & \shortstack{Window\\16} \\
    \midrule
    Hyena & 
    -8.99\% & 12.30\% & 27.30\% \\
    MambaVision & 
    10.03\% & 34.19\% & 46.61\% \\
    \bottomrule
    \end{tabular}}
\end{table}

We observe speedups with longer context lengths, with both Hyena and MambaVision achieving over 80\% speedups with 4-pixel patches in ViT. At smaller context lengths, we observe the alternative operators slow down training, as expected given the complexity terms (Section \ref{sec:approach}).

\paragraph{Attention Alternatives Enable Longer Context Lengths.} In addition to speeding up training at long context lengths, both Hyena and MambaVision enabled longer context lengths than could be achieved with self-attention given our hardware. For example, in abdominal CT segmentation, memory limitations prevented a self-attention ViT from being trained with 8-pixel patches, while both Hyena and MambaVision reduced memory requirements enough to train with 8-pixel patches. This enabled Hyena and/or MambaVision to exceed the maximum performance achieved by attention-based ViTs on multiple tasks, including vessel segmentation, organ segmentation, microscopy denoising, and pneumothorax classification.

\section{Discussion and Conclusion}

In this study, we evaluated the impact of context length on the performance and efficiency of transformers for biomedical image analysis.
We further investigated two alternatives to self-attention---Hyena and MambaVision---on diverse imaging tasks.

\paragraph{Key Findings.}
Our results indicate a strong relationship between patch size and task performance, particularly for pixel-level prediction tasks. Smaller patch sizes, which correspond to longer context lengths, consistently yielded better performance. This finding underscores the importance of preserving high-resolution information in biomedical images, which often contain critical fine-grained details necessary for accurate predictions. 

In contrast, Swin's window size did not strongly impact performance, although denoising tasks showed some performance gains with larger windows. This suggests that while local context is crucial, Swin's hierarchical design may already provide a sufficient balance between local and global information for many tasks. In this case, dedicating more context length to preserving image resolution may be more impactful than extending context length to achieve larger attention windows.

We found both Hyena and MambaVision to be promising alternatives to self-attention that enable smaller patches and greater attention windows. 
For ViT pixel-level prediction tasks, we found that both operators could exceed the performance achieved by self-attention networks while also offering significant speedups---up to 80\% faster---for longer context lengths.  This efficiency gain is critical for biomedical applications, where high-resolution images are common and computational resources are often a limiting factor in network design.

\paragraph{Limitations and Future Work.}
This work focuses on a specific set of alternative operators. Further work may explore a wider range of efficient attention alternatives and their suitability for diverse medical imaging tasks. 
Additionally, the datasets we used are relatively small. Future work using larger datasets may show additional strengths and weaknesses of each of these operators. Similarly, the maximum context lengths in this work were limited by GPU memory. Future work may further extend context length with alternative training environments. Finally, future work may study how context length and attention alternatives impact pretraining strategies and self-supervision performance.

\paragraph{Conclusion.}

In this study, we explored the role that context length plays in biomedical image analysis, investigating the relationship between context length, performance, and efficiency. 
We found that smaller patch sizes improved performance across a range of task and data types, underscoring the importance of preserving high-resolution information in biomedical image analysis. 
However, the increased computational demands associated with longer context lengths present challenges for practical clinical applications.

We demonstrated that replacing the traditional attention operator with alternatives like Hyena or Mamba can help alleviate these computational challenges. These operators facilitate computation over longer context lengths by reducing the compute time and memory requirements while maintaining---sometimes even improving---performance, particularly for pixel-level prediction tasks. The efficiency of Hyena and Mamba offers advantages for real-time, real-world clinical implementations, where computational resources can be limited, fast processing is desired, and performance is paramount.

In conclusion, our findings can inform the design of model backbones for biomedical imaging tasks and provide insights for the development of new biomedical imaging models that balance performance and efficiency, ultimately supporting more effective solutions for biomedical image analysis.

\newpage

\bibliography{references}

\begin{thebibliography}{65}
\providecommand{\natexlab}[1]{#1}
\providecommand{\url}[1]{\texttt{#1}}
\expandafter\ifx\csname urlstyle\endcsname\relax
  \providecommand{\doi}[1]{doi: #1}\else
  \providecommand{\doi}{doi: \begingroup \urlstyle{rm}\Url}\fi

\bibitem[Antonelli et~al.(2022)Antonelli, Reinke, Bakas, Farahani, Kopp-Schneider, Landman, Litjens, Menze, Ronneberger, Summers, et~al.]{antonelli2022medical}
Michela Antonelli, Annika Reinke, Spyridon Bakas, Keyvan Farahani, Annette Kopp-Schneider, Bennett~A Landman, Geert Litjens, Bjoern Menze, Olaf Ronneberger, Ronald~M Summers, et~al.
\newblock The medical segmentation decathlon.
\newblock \emph{Nature communications}, 13\penalty0 (1):\penalty0 4128, 2022.

\bibitem[Archit and Pape(2024)]{archit2024vim}
Anwai Archit and Constantin Pape.
\newblock Vim-unet: Vision mamba for biomedical segmentation.
\newblock \emph{arXiv preprint arXiv:2404.07705}, 2024.

\bibitem[Arora et~al.(2023)Arora, Eyuboglu, Timalsina, Johnson, Poli, Zou, Rudra, and Ré]{zoology2023}
Simran Arora, Sabri Eyuboglu, Aman Timalsina, Isys Johnson, Michael Poli, James Zou, Atri Rudra, and Christopher Ré.
\newblock Zoology: Measuring and improving recall in efficient language models.
\newblock \emph{arXiv:2312.04927}, 2023.

\bibitem[Beltagy et~al.(2020)Beltagy, Peters, and Cohan]{beltagy2020longformer}
Iz~Beltagy, Matthew~E Peters, and Arman Cohan.
\newblock Longformer: The long-document transformer.
\newblock \emph{arXiv preprint arXiv:2004.05150}, 2020.

\bibitem[Berisha et~al.(2021)Berisha, Krantsevich, Hahn, Hahn, Dasarathy, Turaga, and Liss]{berisha2021digital}
Visar Berisha, Chelsea Krantsevich, P~Richard Hahn, Shira Hahn, Gautam Dasarathy, Pavan Turaga, and Julie Liss.
\newblock Digital medicine and the curse of dimensionality.
\newblock \emph{NPJ digital medicine}, 4\penalty0 (1):\penalty0 153, 2021.

\bibitem[Beyer et~al.(2023)Beyer, Izmailov, Kolesnikov, Caron, Kornblith, Zhai, Minderer, Tschannen, Alabdulmohsin, and Pavetic]{beyer2023flexivit}
Lucas Beyer, Pavel Izmailov, Alexander Kolesnikov, Mathilde Caron, Simon Kornblith, Xiaohua Zhai, Matthias Minderer, Michael Tschannen, Ibrahim Alabdulmohsin, and Filip Pavetic.
\newblock Flexivit: One model for all patch sizes.
\newblock In \emph{Proceedings of the IEEE/CVF Conference on Computer Vision and Pattern Recognition}, pages 14496--14506, 2023.

\bibitem[Cardoso et~al.(2022)Cardoso, Li, Brown, Ma, Kerfoot, Wang, Murrey, Myronenko, Zhao, Yang, Nath, He, Xu, Hatamizadeh, Myronenko, Zhu, Liu, Zheng, Tang, Yang, Zephyr, Hashemian, Alle, Darestani, Budd, Modat, Vercauteren, Wang, Li, Hu, Fu, Gorman, Johnson, Genereaux, Erdal, Gupta, Diaz-Pinto, Dourson, Maier-Hein, Jaeger, Baumgartner, Kalpathy-Cramer, Flores, Kirby, Cooper, Roth, Xu, Bericat, Floca, Zhou, Shuaib, Farahani, Maier-Hein, Aylward, Dogra, Ourselin, and Feng]{cardoso2022monaiopensourceframeworkdeep}
M.~Jorge Cardoso, Wenqi Li, Richard Brown, Nic Ma, Eric Kerfoot, Yiheng Wang, Benjamin Murrey, Andriy Myronenko, Can Zhao, Dong Yang, Vishwesh Nath, Yufan He, Ziyue Xu, Ali Hatamizadeh, Andriy Myronenko, Wentao Zhu, Yun Liu, Mingxin Zheng, Yucheng Tang, Isaac Yang, Michael Zephyr, Behrooz Hashemian, Sachidanand Alle, Mohammad~Zalbagi Darestani, Charlie Budd, Marc Modat, Tom Vercauteren, Guotai Wang, Yiwen Li, Yipeng Hu, Yunguan Fu, Benjamin Gorman, Hans Johnson, Brad Genereaux, Barbaros~S. Erdal, Vikash Gupta, Andres Diaz-Pinto, Andre Dourson, Lena Maier-Hein, Paul~F. Jaeger, Michael Baumgartner, Jayashree Kalpathy-Cramer, Mona Flores, Justin Kirby, Lee A.~D. Cooper, Holger~R. Roth, Daguang Xu, David Bericat, Ralf Floca, S.~Kevin Zhou, Haris Shuaib, Keyvan Farahani, Klaus~H. Maier-Hein, Stephen Aylward, Prerna Dogra, Sebastien Ourselin, and Andrew Feng.
\newblock Monai: An open-source framework for deep learning in healthcare, 2022.

\bibitem[Child et~al.(2019)Child, Gray, Radford, and Sutskever]{child2019generating}
Rewon Child, Scott Gray, Alec Radford, and Ilya Sutskever.
\newblock Generating long sequences with sparse transformers.
\newblock \emph{arXiv preprint arXiv:1904.10509}, 2019.

\bibitem[Choromanski et~al.(2020)Choromanski, Likhosherstov, Dohan, Song, Gane, Sarlos, Hawkins, Davis, Mohiuddin, Kaiser, et~al.]{choromanski2020rethinking}
Krzysztof Choromanski, Valerii Likhosherstov, David Dohan, Xingyou Song, Andreea Gane, Tamas Sarlos, Peter Hawkins, Jared Davis, Afroz Mohiuddin, Lukasz Kaiser, et~al.
\newblock Rethinking attention with performers.
\newblock \emph{arXiv preprint arXiv:2009.14794}, 2020.

\bibitem[Colak et~al.(2021)Colak, Kitamura, Hobbs, Wu, Lungren, Prevedello, Kalpathy-Cramer, Ball, Shih, Stein, et~al.]{colak2021rsna}
Errol Colak, Felipe~C Kitamura, Stephen~B Hobbs, Carol~C Wu, Matthew~P Lungren, Luciano~M Prevedello, Jayashree Kalpathy-Cramer, Robyn~L Ball, George Shih, Anouk Stein, et~al.
\newblock The rsna pulmonary embolism ct dataset.
\newblock \emph{Radiology: Artificial Intelligence}, 3\penalty0 (2):\penalty0 e200254, 2021.

\bibitem[Dao(2023)]{dao2023flashattention}
Tri Dao.
\newblock Flashattention-2: Faster attention with better parallelism and work partitioning.
\newblock \emph{arXiv preprint arXiv:2307.08691}, 2023.

\bibitem[Dao et~al.(2022)Dao, Fu, Ermon, Rudra, and R{\'e}]{dao2022flashattention}
Tri Dao, Dan Fu, Stefano Ermon, Atri Rudra, and Christopher R{\'e}.
\newblock Flashattention: Fast and memory-efficient exact attention with io-awareness.
\newblock \emph{Advances in Neural Information Processing Systems}, 35:\penalty0 16344--16359, 2022.

\bibitem[Ding et~al.(2022)Ding, Zhang, Han, and Ding]{ding2022scaling}
Xiaohan Ding, Xiangyu Zhang, Jungong Han, and Guiguang Ding.
\newblock Scaling up your kernels to 31x31: Revisiting large kernel design in cnns.
\newblock In \emph{Proceedings of the IEEE/CVF conference on computer vision and pattern recognition}, pages 11963--11975, 2022.

\bibitem[Dinsdale et~al.(2022)Dinsdale, Bluemke, Sundaresan, Jenkinson, Smith, and Namburete]{dinsdale2022challenges}
Nicola~K Dinsdale, Emma Bluemke, Vaanathi Sundaresan, Mark Jenkinson, Stephen~M Smith, and Ana~IL Namburete.
\newblock Challenges for machine learning in clinical translation of big data imaging studies.
\newblock \emph{Neuron}, 110\penalty0 (23):\penalty0 3866--3881, 2022.

\bibitem[Dong et~al.(2023)Dong, Tang, Li, and Zhao]{dong2023survey}
Zican Dong, Tianyi Tang, Lunyi Li, and Wayne~Xin Zhao.
\newblock A survey on long text modeling with transformers.
\newblock \emph{arXiv preprint arXiv:2302.14502}, 2023.

\bibitem[Dosovitskiy et~al.(2020)Dosovitskiy, Beyer, Kolesnikov, Weissenborn, Zhai, Unterthiner, Dehghani, Minderer, Heigold, Gelly, et~al.]{dosovitskiy2020image}
Alexey Dosovitskiy, Lucas Beyer, Alexander Kolesnikov, Dirk Weissenborn, Xiaohua Zhai, Thomas Unterthiner, Mostafa Dehghani, Matthias Minderer, Georg Heigold, Sylvain Gelly, et~al.
\newblock An image is worth 16x16 words: Transformers for image recognition at scale.
\newblock \emph{arXiv preprint arXiv:2010.11929}, 2020.

\bibitem[Feng et~al.(2021)Feng, Azzollini, Kim, Jin, Gordon, Yeoh, Kim, Han, Lee, Patel, et~al.]{feng2021curation}
Sijing Feng, Damian Azzollini, Ji~Soo Kim, Cheng-Kai Jin, Simon~P Gordon, Jason Yeoh, Eve Kim, Mina Han, Andrew Lee, Aakash Patel, et~al.
\newblock Curation of the candid-ptx dataset with free-text reports.
\newblock \emph{Radiology: Artificial Intelligence}, 3\penalty0 (6):\penalty0 e210136, 2021.

\bibitem[Fillioux et~al.(2023)Fillioux, Boyd, Vakalopoulou, Courn{\`e}de, and Christodoulidis]{fillioux2023structured}
Leo Fillioux, Joseph Boyd, Maria Vakalopoulou, Paul-Henry Courn{\`e}de, and Stergios Christodoulidis.
\newblock Structured state space models for multiple instance learning in digital pathology.
\newblock In \emph{International Conference on Medical Image Computing and Computer-Assisted Intervention}, pages 594--604. Springer, 2023.

\bibitem[Fu et~al.(2022)Fu, Dao, Saab, Thomas, Rudra, and R{\'e}]{fu2022hungry}
Daniel~Y Fu, Tri Dao, Khaled~K Saab, Armin~W Thomas, Atri Rudra, and Christopher R{\'e}.
\newblock Hungry hungry hippos: Towards language modeling with state space models.
\newblock \emph{arXiv preprint arXiv:2212.14052}, 2022.

\bibitem[Gu and Dao(2023)]{gu2023mamba}
Albert Gu and Tri Dao.
\newblock Mamba: Linear-time sequence modeling with selective state spaces.
\newblock \emph{arXiv preprint arXiv:2312.00752}, 2023.

\bibitem[Gu et~al.(2021{\natexlab{a}})Gu, Goel, and R{\'e}]{gu2021efficiently}
Albert Gu, Karan Goel, and Christopher R{\'e}.
\newblock Efficiently modeling long sequences with structured state spaces.
\newblock \emph{arXiv preprint arXiv:2111.00396}, 2021{\natexlab{a}}.

\bibitem[Gu et~al.(2021{\natexlab{b}})Gu, Johnson, Goel, Saab, Dao, Rudra, and R{\'e}]{gu2021combining}
Albert Gu, Isys Johnson, Karan Goel, Khaled Saab, Tri Dao, Atri Rudra, and Christopher R{\'e}.
\newblock Combining recurrent, convolutional, and continuous-time models with linear state space layers.
\newblock \emph{Advances in neural information processing systems}, 34:\penalty0 572--585, 2021{\natexlab{b}}.

\bibitem[Han et~al.(2022)Han, Wang, Chen, Chen, Guo, Liu, Tang, Xiao, Xu, Xu, et~al.]{han2022survey}
Kai Han, Yunhe Wang, Hanting Chen, Xinghao Chen, Jianyuan Guo, Zhenhua Liu, Yehui Tang, An~Xiao, Chunjing Xu, Yixing Xu, et~al.
\newblock A survey on vision transformer.
\newblock \emph{IEEE transactions on pattern analysis and machine intelligence}, 45\penalty0 (1):\penalty0 87--110, 2022.

\bibitem[Hatamizadeh and Kautz(2024)]{hatamizadeh2024mambavision}
Ali Hatamizadeh and Jan Kautz.
\newblock Mambavision: A hybrid mamba-transformer vision backbone.
\newblock \emph{arXiv preprint arXiv:2407.08083}, 2024.

\bibitem[Hatamizadeh et~al.(2022)Hatamizadeh, Tang, Nath, Yang, Myronenko, Landman, Roth, and Xu]{hatamizadeh2022unetr}
Ali Hatamizadeh, Yucheng Tang, Vishwesh Nath, Dong Yang, Andriy Myronenko, Bennett Landman, Holger~R Roth, and Daguang Xu.
\newblock Unetr: Transformers for 3d medical image segmentation.
\newblock In \emph{Proceedings of the IEEE/CVF winter conference on applications of computer vision}, pages 574--584, 2022.

\bibitem[He et~al.(2023)He, Gan, Li, Rekik, Yin, Ji, Gao, Wang, Zhang, and Shen]{he2023transformers}
Kelei He, Chen Gan, Zhuoyuan Li, Islem Rekik, Zihao Yin, Wen Ji, Yang Gao, Qian Wang, Junfeng Zhang, and Dinggang Shen.
\newblock Transformers in medical image analysis.
\newblock \emph{Intelligent Medicine}, 3\penalty0 (1):\penalty0 59--78, 2023.

\bibitem[Hu et~al.(2022)Hu, Debnath, Xie, and Chen]{hu2022exploring}
Ronghang Hu, Shoubhik Debnath, Saining Xie, and Xinlei Chen.
\newblock Exploring long-sequence masked autoencoders.
\newblock \emph{arXiv preprint arXiv:2210.07224}, 2022.

\bibitem[Huang et~al.(2023)Huang, Xu, Jiang, Lai, Li, Yao, Chen, Yang, Xin, and Ma]{huang2023advancing}
Yunpeng Huang, Jingwei Xu, Zixu Jiang, Junyu Lai, Zenan Li, Yuan Yao, Taolue Chen, Lijuan Yang, Zhou Xin, and Xiaoxing Ma.
\newblock Advancing transformer architecture in long-context large language models: A comprehensive survey.
\newblock \emph{arXiv preprint arXiv:2311.12351}, 2023.

\bibitem[Ibrahimovic(2023)]{ibrahimovic2023optimizing}
E~Ibrahimovic.
\newblock Optimizing vision transformer performance with customizable parameters.
\newblock In \emph{2023 46th MIPRO ICT and Electronics Convention (MIPRO)}, pages 1721--1726. IEEE, 2023.

\bibitem[Jin et~al.(2022)Jin, Huang, Zhou, Li, Yan, Sun, Zhang, Wang, and Ye]{jin2022fives}
Kai Jin, Xingru Huang, Jingxing Zhou, Yunxiang Li, Yan Yan, Yibao Sun, Qianni Zhang, Yaqi Wang, and Juan Ye.
\newblock Fives: A fundus image dataset for artificial intelligence based vessel segmentation.
\newblock \emph{Scientific data}, 9\penalty0 (1):\penalty0 475, 2022.

\bibitem[Katharopoulos et~al.(2020)Katharopoulos, Vyas, Pappas, and Fleuret]{katharopoulos2020transformers}
Angelos Katharopoulos, Apoorv Vyas, Nikolaos Pappas, and Fran{\c{c}}ois Fleuret.
\newblock Transformers are rnns: Fast autoregressive transformers with linear attention.
\newblock In \emph{International conference on machine learning}, pages 5156--5165. PMLR, 2020.

\bibitem[Keles et~al.(2023)Keles, Wijewardena, and Hegde]{keles2023computational}
Feyza~Duman Keles, Pruthuvi~Mahesakya Wijewardena, and Chinmay Hegde.
\newblock On the computational complexity of self-attention.
\newblock In \emph{International Conference on Algorithmic Learning Theory}, pages 597--619. PMLR, 2023.

\bibitem[Khan et~al.(2022)Khan, Naseer, Hayat, Zamir, Khan, and Shah]{khan2022transformers}
Salman Khan, Muzammal Naseer, Munawar Hayat, Syed~Waqas Zamir, Fahad~Shahbaz Khan, and Mubarak Shah.
\newblock Transformers in vision: A survey.
\newblock \emph{ACM computing surveys (CSUR)}, 54\penalty0 (10s):\penalty0 1--41, 2022.

\bibitem[Liu et~al.(2024)Liu, Tian, Zhao, Yu, Xie, Wang, Ye, and Liu]{liu2024vmambavisualstatespace}
Yue Liu, Yunjie Tian, Yuzhong Zhao, Hongtian Yu, Lingxi Xie, Yaowei Wang, Qixiang Ye, and Yunfan Liu.
\newblock Vmamba: Visual state space model, 2024.

\bibitem[Liu et~al.(2021)Liu, Lin, Cao, Hu, Wei, Zhang, Lin, and Guo]{liu2021swin}
Ze~Liu, Yutong Lin, Yue Cao, Han Hu, Yixuan Wei, Zheng Zhang, Stephen Lin, and Baining Guo.
\newblock Swin transformer: Hierarchical vision transformer using shifted windows.
\newblock In \emph{Proceedings of the IEEE/CVF international conference on computer vision}, pages 10012--10022, 2021.

\bibitem[Ma et~al.(2024)Ma, Li, and Wang]{ma2024u}
Jun Ma, Feifei Li, and Bo~Wang.
\newblock U-mamba: Enhancing long-range dependency for biomedical image segmentation.
\newblock \emph{arXiv preprint arXiv:2401.04722}, 2024.

\bibitem[McKinzie et~al.(2024)McKinzie, Gan, Fauconnier, Dodge, Zhang, Dufter, Shah, Du, Peng, Weers, et~al.]{mckinzie2024mm1}
Brandon McKinzie, Zhe Gan, Jean-Philippe Fauconnier, Sam Dodge, Bowen Zhang, Philipp Dufter, Dhruti Shah, Xianzhi Du, Futang Peng, Floris Weers, et~al.
\newblock Mm1: Methods, analysis \& insights from multimodal llm pre-training.
\newblock \emph{arXiv preprint arXiv:2403.09611}, 2024.

\bibitem[Meng et~al.(2024)Meng, Yang, Tian, Dai, Wu, Gao, and Jiang]{meng2024deepstack}
Lingchen Meng, Jianwei Yang, Rui Tian, Xiyang Dai, Zuxuan Wu, Jianfeng Gao, and Yu-Gang Jiang.
\newblock Deepstack: Deeply stacking visual tokens is surprisingly simple and effective for lmms.
\newblock \emph{arXiv preprint arXiv:2406.04334}, 2024.

\bibitem[Nasiri-Sarvi et~al.(2024)Nasiri-Sarvi, Trinh, Rivaz, and Hosseini]{nasiri2024vim4path}
Ali Nasiri-Sarvi, Vincent Quoc-Huy Trinh, Hassan Rivaz, and Mahdi~S Hosseini.
\newblock Vim4path: Self-supervised vision mamba for histopathology images.
\newblock In \emph{Proceedings of the IEEE/CVF Conference on Computer Vision and Pattern Recognition}, pages 6894--6903, 2024.

\bibitem[Nguyen et~al.(2024)Nguyen, Assran, Jain, Oswald, Snoek, and Chen]{nguyen2024image}
Duy-Kien Nguyen, Mahmoud Assran, Unnat Jain, Martin~R Oswald, Cees~GM Snoek, and Xinlei Chen.
\newblock An image is worth more than 16x16 patches: Exploring transformers on individual pixels.
\newblock \emph{arXiv preprint arXiv:2406.09415}, 2024.

\bibitem[Nguyen et~al.(2022)Nguyen, Goel, Gu, Downs, Shah, Dao, Baccus, and R{\'e}]{nguyen2022s4nd}
Eric Nguyen, Karan Goel, Albert Gu, Gordon Downs, Preey Shah, Tri Dao, Stephen Baccus, and Christopher R{\'e}.
\newblock S4nd: Modeling images and videos as multidimensional signals with state spaces.
\newblock \emph{Advances in neural information processing systems}, 35:\penalty0 2846--2861, 2022.

\bibitem[Pawar et~al.(2024)Pawar, Tonmoy, Zaman, Jain, Chadha, and Das]{pawar2024and}
Saurav Pawar, SM~Tonmoy, SM~Zaman, Vinija Jain, Aman Chadha, and Amitava Das.
\newblock The what, why, and how of context length extension techniques in large language models--a detailed survey.
\newblock \emph{arXiv preprint arXiv:2401.07872}, 2024.

\bibitem[Peebles and Xie(2023)]{peebles2023scalable}
William Peebles and Saining Xie.
\newblock Scalable diffusion models with transformers.
\newblock In \emph{Proceedings of the IEEE/CVF International Conference on Computer Vision}, pages 4195--4205, 2023.

\bibitem[Pei et~al.(2024)Pei, Huang, and Xu]{pei2024efficientvmamba}
Xiaohuan Pei, Tao Huang, and Chang Xu.
\newblock Efficientvmamba: Atrous selective scan for light weight visual mamba.
\newblock \emph{arXiv preprint arXiv:2403.09977}, 2024.

\bibitem[Peng et~al.(2023)Peng, Alcaide, Anthony, Albalak, Arcadinho, Biderman, Cao, Cheng, Chung, Grella, et~al.]{peng2023rwkv}
Bo~Peng, Eric Alcaide, Quentin Anthony, Alon Albalak, Samuel Arcadinho, Stella Biderman, Huanqi Cao, Xin Cheng, Michael Chung, Matteo Grella, et~al.
\newblock Rwkv: Reinventing rnns for the transformer era.
\newblock \emph{arXiv preprint arXiv:2305.13048}, 2023.

\bibitem[Poli et~al.(2023)Poli, Massaroli, Nguyen, Fu, Dao, Baccus, Bengio, Ermon, and R{\'e}]{poli2023hyena}
Michael Poli, Stefano Massaroli, Eric Nguyen, Daniel~Y Fu, Tri Dao, Stephen Baccus, Yoshua Bengio, Stefano Ermon, and Christopher R{\'e}.
\newblock Hyena hierarchy: Towards larger convolutional language models.
\newblock In \emph{International Conference on Machine Learning}, pages 28043--28078. PMLR, 2023.

\bibitem[Qu et~al.(2024)Qu, Zhang, Qiao, Tang, Yuille, Zhou, et~al.]{qu2024abdomenatlas}
Chongyu Qu, Tiezheng Zhang, Hualin Qiao, Yucheng Tang, Alan~L Yuille, Zongwei Zhou, et~al.
\newblock Abdomenatlas-8k: Annotating 8,000 ct volumes for multi-organ segmentation in three weeks.
\newblock \emph{Advances in Neural Information Processing Systems}, 36, 2024.

\bibitem[Sabottke and Spieler(2020)]{sabottke2020effect}
Carl~F Sabottke and Bradley~M Spieler.
\newblock The effect of image resolution on deep learning in radiography.
\newblock \emph{Radiology: Artificial Intelligence}, 2\penalty0 (1):\penalty0 e190015, 2020.

\bibitem[Shah et~al.(2024)Shah, Bikshandi, Zhang, Thakkar, Ramani, and Dao]{shah2024flashattention}
Jay Shah, Ganesh Bikshandi, Ying Zhang, Vijay Thakkar, Pradeep Ramani, and Tri Dao.
\newblock Flashattention-3: Fast and accurate attention with asynchrony and low-precision.
\newblock \emph{arXiv preprint arXiv:2407.08608}, 2024.

\bibitem[Shamshad et~al.(2023)Shamshad, Khan, Zamir, Khan, Hayat, Khan, and Fu]{shamshad2023transformers}
Fahad Shamshad, Salman Khan, Syed~Waqas Zamir, Muhammad~Haris Khan, Munawar Hayat, Fahad~Shahbaz Khan, and Huazhu Fu.
\newblock Transformers in medical imaging: A survey.
\newblock \emph{Medical Image Analysis}, 88:\penalty0 102802, 2023.

\bibitem[Sun et~al.(2023)Sun, Dong, Huang, Ma, Xia, Xue, Wang, and Wei]{sun2023retentive}
Yutao Sun, Li~Dong, Shaohan Huang, Shuming Ma, Yuqing Xia, Jilong Xue, Jianyong Wang, and Furu Wei.
\newblock Retentive network: A successor to transformer for large language models.
\newblock \emph{arXiv preprint arXiv:2307.08621}, 2023.

\bibitem[Suzuki(2017)]{suzuki2017overview}
Kenji Suzuki.
\newblock Overview of deep learning in medical imaging.
\newblock \emph{Radiological physics and technology}, 10\penalty0 (3):\penalty0 257--273, 2017.

\bibitem[Tay et~al.(2020)Tay, Dehghani, Bahri, and Metzler]{DBLP:journals/corr/abs-2009-06732}
Yi~Tay, Mostafa Dehghani, Dara Bahri, and Donald Metzler.
\newblock Efficient transformers: {A} survey.
\newblock \emph{CoRR}, abs/2009.06732, 2020.

\bibitem[Thambawita et~al.(2021)Thambawita, Str{\"u}mke, Hicks, Halvorsen, Parasa, and Riegler]{thambawita2021impact}
Vajira Thambawita, Inga Str{\"u}mke, Steven~A Hicks, P{\aa}l Halvorsen, Sravanthi Parasa, and Michael~A Riegler.
\newblock Impact of image resolution on deep learning performance in endoscopy image classification: An experimental study using a large dataset of endoscopic images.
\newblock \emph{Diagnostics}, 11\penalty0 (12):\penalty0 2183, 2021.

\bibitem[Than et~al.(2021)Than, Thon, Rijal, Kassim, Yunus, Noor, and Then]{than2021preliminary}
Joel~CM Than, Pun~Liang Thon, Omar~Mohd Rijal, Rosminah~M Kassim, Ashari Yunus, Norliza~M Noor, and Patrick Then.
\newblock Preliminary study on patch sizes in vision transformers (vit) for covid-19 and diseased lungs classification.
\newblock In \emph{2021 IEEE National Biomedical Engineering Conference (NBEC)}, pages 146--150. IEEE, 2021.

\bibitem[Touvron et~al.(2021)Touvron, Cord, Douze, Massa, Sablayrolles, and J{\'e}gou]{touvron2021training}
Hugo Touvron, Matthieu Cord, Matthijs Douze, Francisco Massa, Alexandre Sablayrolles, and Herv{\'e} J{\'e}gou.
\newblock Training data-efficient image transformers \& distillation through attention.
\newblock In \emph{International conference on machine learning}, pages 10347--10357. PMLR, 2021.

\bibitem[Tsirmpas et~al.(2024)Tsirmpas, Gkionis, Papadopoulos, and Mademlis]{tsirmpas2024neural}
Dimitrios Tsirmpas, Ioannis Gkionis, Georgios~Th Papadopoulos, and Ioannis Mademlis.
\newblock Neural natural language processing for long texts: A survey on classification and summarization.
\newblock \emph{Engineering Applications of Artificial Intelligence}, 133:\penalty0 108231, 2024.

\bibitem[Vaswani(2017)]{vaswani2017attention}
Ashish Vaswani.
\newblock Attention is all you need.
\newblock \emph{arXiv preprint arXiv:1706.03762}, 2017.

\bibitem[Wang et~al.(2021)Wang, Xie, Li, Fan, Song, Liang, Lu, Luo, and Shao]{wang2021pyramid}
Wenhai Wang, Enze Xie, Xiang Li, Deng-Ping Fan, Kaitao Song, Ding Liang, Tong Lu, Ping Luo, and Ling Shao.
\newblock Pyramid vision transformer: A versatile backbone for dense prediction without convolutions.
\newblock In \emph{Proceedings of the IEEE/CVF international conference on computer vision}, pages 568--578, 2021.

\bibitem[Wang et~al.(2024)Wang, Zheng, Zhang, Cui, and Li]{wang2024mamba}
Ziyang Wang, Jian-Qing Zheng, Yichi Zhang, Ge~Cui, and Lei Li.
\newblock Mamba-unet: Unet-like pure visual mamba for medical image segmentation.
\newblock \emph{arXiv preprint arXiv:2402.05079}, 2024.

\bibitem[Xiao et~al.(2018)Xiao, Liu, Zhou, Jiang, and Sun]{xiao2018unified}
Tete Xiao, Yingcheng Liu, Bolei Zhou, Yuning Jiang, and Jian Sun.
\newblock Unified perceptual parsing for scene understanding.
\newblock In \emph{Proceedings of the European conference on computer vision (ECCV)}, pages 418--434, 2018.

\bibitem[Xie et~al.(2021)Xie, Wang, Yu, Anandkumar, Alvarez, and Luo]{xie2021segformer}
Enze Xie, Wenhai Wang, Zhiding Yu, Anima Anandkumar, Jose~M Alvarez, and Ping Luo.
\newblock Segformer: Simple and efficient design for semantic segmentation with transformers.
\newblock \emph{Advances in neural information processing systems}, 34:\penalty0 12077--12090, 2021.

\bibitem[Xing et~al.(2024)Xing, Ye, Yang, Liu, and Zhu]{xing2024segmamba}
Zhaohu Xing, Tian Ye, Yijun Yang, Guang Liu, and Lei Zhu.
\newblock Segmamba: Long-range sequential modeling mamba for 3d medical image segmentation.
\newblock \emph{arXiv preprint arXiv:2401.13560}, 2024.

\bibitem[Zhou et~al.(2020)Zhou, El~Helou, Sage, Laroche, Seitz, and S{\"u}sstrunk]{zhou2020w2s}
Ruofan Zhou, Majed El~Helou, Daniel Sage, Thierry Laroche, Arne Seitz, and Sabine S{\"u}sstrunk.
\newblock W2s: microscopy data with joint denoising and super-resolution for widefield to sim mapping.
\newblock In \emph{Computer Vision--ECCV 2020 Workshops: Glasgow, UK, August 23--28, 2020, Proceedings, Part I 16}, pages 474--491. Springer, 2020.

\bibitem[Zhu et~al.(2024)Zhu, Liao, Zhang, Wang, Liu, and Wang]{zhu2024vision}
Lianghui Zhu, Bencheng Liao, Qian Zhang, Xinlong Wang, Wenyu Liu, and Xinggang Wang.
\newblock Vision mamba: Efficient visual representation learning with bidirectional state space model.
\newblock \emph{arXiv preprint arXiv:2401.09417}, 2024.

\end{thebibliography}

\newpage 

\appendix

\section{Training Details}
\label{apd:training details}

\subsection{Hyperparameters}

\begin{table*}[hbtp]
\floatconts
  {tab:hyperparams_vit}
  {\caption{Selected learning rates for the ViT backbone.}}
    {
    \centering
    \resizebox{\textwidth}{!}{
    \begin{tabular}{c|cccc|cccc|cccc}
    \toprule
     & \multicolumn{4}{c|}{ViT with Attention} 
     & \multicolumn{4}{c|}{ViT with Hyena} 
     & \multicolumn{4}{c}{ViT with MambaVision} \\
     & Patch 4 & Patch 8 & Patch 16 & Patch 32 
     & Patch 4 & Patch 8 & Patch 16 & Patch 32 
     & Patch 4 & Patch 8 & Patch 16 & Patch 32 \\
    \midrule
    Vessel & 
    X & X & X & 1e-3 & 
    X & X & 1e-3 & 1e-3 & 
    X & X & 1e-3 & 1e-3 \\
    Ab. CT & 
    X & X & 1e-3 & 1e-3 & 
    X & 1e-3 & 1e-3 & 1e-3 & 
    X & 1e-3 & 1e-3 & 1e-3 \\
    Microscopy & 
    X & X & 1e-3 & 1e-3 & 
    X & 1e-3 & 1e-3 & 1e-3 & 
    1e-3 & 1e-3 & 1e-3 & 1e-3 \\
    CMR & 
    1e-3 & 1e-2 & 1e-2 & 1e-2 & 
    1e-3 & 1e-3 & 1e-2 & 1e-2 & 
    1e-3 & 1e-3 & 1e-3 & 1e-3 \\
    Pneumothorax & 
    X & X & 1e-4 & 1e-4 & 
    X & 1e-4 & 1e-4 & 1e-4 & 
    X & 1e-4 & 1e-4 & 1e-4 \\ 
    Embolism & 
    X & 1e-5 & 1e-4 & 1e-4 & 
    1e-3 & 1e-3 & 1e-5 & 1e-5 &
    1e-5 & 1e-5 & 1e-5 & 1e-5 \\
    \bottomrule
    \end{tabular}}}
\end{table*}

\begin{table*}[hbtp]
\floatconts
  {tab:hyperparams_swin}
  {\caption{Selected learning rates for the Swin backbone.}}
    {
    \centering
    \resizebox{.94\textwidth}{!}{
    \begin{tabular}{c|ccc|ccc|ccc}
    \toprule
     & \multicolumn{3}{c|}{Swin with Attention} 
     & \multicolumn{3}{c|}{Swin with Hyena} 
     & \multicolumn{3}{c}{Swin with MambaVision} \\
     & Window 16 & Window 8 & Window 4
     & Window 16 & Window 8 & Window 4
     & Window 16 & Window 8 & Window 4\\
    \midrule
    Vessel & 
    1e-3 & 1e-3 & 1e-3 & 
    1e-3 & 1e-3 & 1e-3 & 
    1e-3 & 1e-3 & 1e-3 \\
    Ab. CT & 
    X & 1e-4 & 1e-4 & 
    1e-3 & 1e-3 & 1e-3 & 
    1e-4 & 1e-4 & 1e-4 \\
    Microscopy & 
    1e-4 & 1e-4 & 1e-4 & 
    1e-4 & 1e-4 & 1e-4 & 
    1e-5 & 1e-4 & 1e-5 \\
    CMR & 
    1e-4 & 1e-4 & 1e-4 & 
    1e-5 & 1e-5 & 1e-5 & 
    1e-4 & 1e-4 & 1e-4 \\
    Pneumothorax & 
    1e-5 & 1e-5 & 1e-5 & 
    1e-5 & 1e-4 & 1e-5 & 
    1e-5 & 1e-5 & 1e-5 \\ 
    Embolism & 
    X & 1e-5 & 1e-5 & 
    1e-5 & 1e-5 & 1e-5 &
    1e-5 & 1e-5 & 1e-5 \\
    \bottomrule
    \end{tabular}}}
\end{table*}

We tuned the learning rate for each experiment from $\{1e-5, 1e-4, 1e-3, 1e-2\}.$ Selected learning rates are given in Table \ref{tab:hyperparams_vit} and Table \ref{tab:hyperparams_swin}. We set batch size to maximize GPU memory. We required a minimum batch size of two to fit on the GPU to enable batch normalization layers. 

\subsection{Data Preprocessing}

For the retinal vessel segmentation dataset \citep{jin2022fives}, we directly used the public data with no additional preprocessing. When training the Swin models, we resized the images to $1024 \times 1024$ to fit onto the GPU.

For the abdominal CT organ segmentation dataset, we used the images supplied by \cite{antonelli2022medical} and segmentation masks supplied by \cite{qu2024abdomenatlas} for the aorta, gall bladder, kidneys, liver, pancreas, postcava, spleen, and stomach. We windowed the CT with a window level of 50 and window width of 400. We resized each axial image using linear interpolation to $256 \times 256$ and center cropped to $64$ axial slices.

For the microscopy denoising dataset \citep{zhou2020w2s}, we treated each of the three supplied channels in the public dataset as different images. We selected a single frame from the widefield images as our low-SNR image and normalized each to zero mean and unit variance. We used the structured-illumination microscopy image as our paired high-SNR image, and scaled the high-SNR image using a least squares fit. 

For the cardiac MR denoising dataset, we used images reconstructed in SNR units, meaning the amplitude of the signal in the reconstructed images is representative of its SNR. We added realistic MRI noise using an MRI noise model, reducing the SNR by a ratio selected from a uniform distribution between $[1,40]$. We center cropped each cine to $128 \times 128$ pixels and 32 frames.

For the pneumothorax dataset \citep{feng2021curation}, we normalized each image between $[0,1]$.

For the pulmonary embolism dataset \citep{colak2021rsna}, we windowed the CT with a window level of 100 and window width of 700. We cropped around the lung region then resized each axial slice to $256 \times 256$ and center cropped the axial slices to $64$ slices, ensuring the embolism was captured in the cropped region.

\subsection{Model Implementation}
We used the ViT and Swin implementations from Monai \citep{cardoso2022monaiopensourceframeworkdeep}.
We used the MambaVision implementation provided by the authors of the MambaVision paper \citep{hatamizadeh2024mambavision}, which calls code provided by the authors of the original Mamba paper \citep{gu2023mamba}.
We used the Hyena implementation from a study on efficient language models \citep{zoology2023}, which provides a simple implementation of the method proposed in the Hyena paper \citep{poli2023hyena}.

\subsection{Model Parameter Count}

As discussed in Section \ref{sec:approach}, changing the patch size in ViT and local attention window in Swin changes the initial patch embedding parameters and task head parameters; otherwise, the backbone parameterization is largely unchanged. We report the number of parameters in the model for each experiment in Tables \ref{tab:vit_parameter_counts} and \ref{tab:swin_parameter_counts}. An X in these tables indicates the configuration could not be run due hardware constraints.

\begin{table*}[hbtp]
\floatconts
  {tab:vit_parameter_counts}
  {\caption{ViT parameter counts in the model backbone/task heads. }}
    {
    \resizebox{.85\textwidth}{!}{
    \centering
    \begin{tabular}{cc|cccc}
    \toprule
     &
     & \shortstack{Patch\\32}
     & \shortstack{Patch\\16}
     & \shortstack{Patch\\8}
     & \shortstack{Patch\\4} \\
    \midrule
    
    \multirow{3}{*}{Vessel}
    & Attn & 24,033,408/4,353,026 & X & X & X \\
    & Hyena & 26,659,776/4,353,026 & 30,493,632/4,328,450 & X & X \\
    & MambaVision & 20,674,176/4,353,026 & 24,508,032/4,328,450 & X & X \\
    \midrule

    \multirow{3}{*}{Ab. CT}
    & Attn & 33,912,960/11,398,346 & 23,246,976/11,283,658 & X & X \\
    & Hyena & 36,539,328/11,398,346 & 25,873,344/11,283,658 & 27,249,600/11,269,322 & X \\
    & MambaVision & 30,553,728/11,398,346 & 19,887,744/11,283,658 & 21,264,000/11,269,322 & X \\
    \midrule

    \multirow{3}{*}{Microscopy}
    & Attn & 22,067,328/4,352,353 & 22,952,064/4,327,777 & X & X \\
    & Hyena & 24,693,696/4,352,353 & 25,578,432/4,327,777 & 30,223,296/4,321,633 & X \\
    & MambaVision & 18,708,096/4,352,353 & 19,592,832/4,327,777 & 24,237,696/4,321,633 & 43,093,632/4,206,945 \\
    \midrule

    \multirow{3}{*}{CMR}
    & Attn & 46,452,864/11,398,945 & 24,475,776/11,284,257 & 22,067,328/11,269,921 & 24,475,776/10,958,625 \\
    & Hyena & 49,079,232/11,398,945 & 27,102,144/11,284,257 & 24,693,696/11,269,921 & 27,102,144/10,958,625 \\
    & MambaVision & 43,093,632/11,398,945 & 21,116,544/11,284,257 & 18,708,096/11,269,921 & 21,116,544/10,958,625 \\
    \midrule

    \multirow{3}{*}{Pneumothorax}
    & Attn & 22,067,712/770 & 22,952,448/770 & X & X \\
    & Hyena & 24,693,696/770 & 25,578,432/770 & 30,223,296/770 & X \\
    & MambaVision & 18,708,096/770 & 19,592,832/770 & 24,237,696/770 & X \\
    \midrule

    \multirow{3}{*}{Embolism}
    & Attn & 33,913,344/770 & 23,247,360/770 & 24,623,616/770 & X \\
    & Hyena & 36,539,328/770 & 25,873,344/770 & 27,249,600/770 & 49,097,664/770 \\
    & MaMambaVisionmba & 30,553,728/770 & 19,887,744/770 & 21,264,000/770 & 43,112,064/770 \\

    \bottomrule
    \end{tabular}}
    }
\end{table*}

\begin{table*}[hbtp]
\floatconts
  {tab:swin_parameter_counts}
  {\caption{Swin parameter counts in the model backbone/task heads.}}
    {
    \resizebox{.7\textwidth}{!}{
    \centering
    \begin{tabular}{cc|ccc}
    \toprule
     &
     & \shortstack{Window\\4}
     & \shortstack{Window\\8}
     & \shortstack{Window\\16} \\
    \midrule
    
    \multirow{3}{*}{Vessel}
    & Attn & 32,222,346/9,263,618 & 32,246,634/9,263,618 & 32,348,202/9,263,618 \\
    & Hyena & 34,799,712/9,263,618 & 34,799,712/9,263,618 & 34,799,712/9,263,618 \\
    & MambaVision & 28,090,272/9,263,618 & 28,090,272/9,263,618 & 28,090,272/9,263,618 \\
    \midrule

    \multirow{3}{*}{Ab. CT}
    & Attn & 38,540,934/12,629,770 & 38,959,350/12,629,770 & X \\
    & Hyena & 41,077,728/12,629,770 & 41,077,728/12,629,770 & 41,077,728/12,629,770 \\
    & MambaVision & 34,368,288/12,629,770 & 34,368,288/12,629,770 & 34,368,288/12,629,770 \\
    \midrule

    \multirow{3}{*}{Microscopy}
    & Attn & 32,221,578/9,261,889 & 32,245,866/9,261,889 & 32,347,434/9,261,889 \\
    & Hyena & 34,798,944/9,261,889 & 34,798,944/9,261,889 & 34,798,944/9,261,889 \\
    & MambaVision & 28,089,504/9,261,889 & 28,089,504/9,261,889 & 28,089,504/9,261,889 \\
    \midrule

    \multirow{3}{*}{CMR}
    & Attn & 38,541,702/12,583,105 & 38,960,118/12,583,105 & 42,605,526/12,583,105 \\
    & Hyena & 41,078,496/12,583,105 & 41,078,496/12,583,105 & 41,078,496/12,583,105 \\
    & MambaVision & 34,369,056/12,583,105 & 34,369,056/12,583,105 & 34,369,056/12,583,105 \\
    \midrule

    \multirow{3}{*}{Pneumothorax}
    & Attn & 32,221,578/3,074 & 32,245,866/3,074 & 32,347,434/3,074 \\
    & Hyena & 34,798,944/3,074 & 34,798,944/3,074 & 34,798,944/3,074 \\
    & MambaVision & 28,089,504/3,074 & 28,089,504/3,074 & 28,089,504/3,074 \\
    \midrule

    \multirow{3}{*}{Embolism}
    & Attn & 38,540,934/3,074 & 38,959,350/3,074 & X \\
    & Hyena & 41,077,728/3,074 & 41,077,728/3,074 & 41,077,728/3,074 \\
    & MambaVision & 34,368,288/3,074 & 34,368,288/3,074 & 34,368,288/3,074 \\

    \bottomrule
    \end{tabular}}
    }
\end{table*}

\section{Additional Results}
\label{apd:additional results}

\subsection{Efficiency}

\subsubsection{Training Timing}

\begin{figure*}[htbp]
\floatconts
  {fig:vit_runtime}
  {\caption{ViT timing. We visualize timing for a forward and backward pass for each task, operator, and patch size. An X on the x-axis indicates that the patch size exceeded our hardware capacity.}}
  {\includegraphics [width=.9\textwidth]{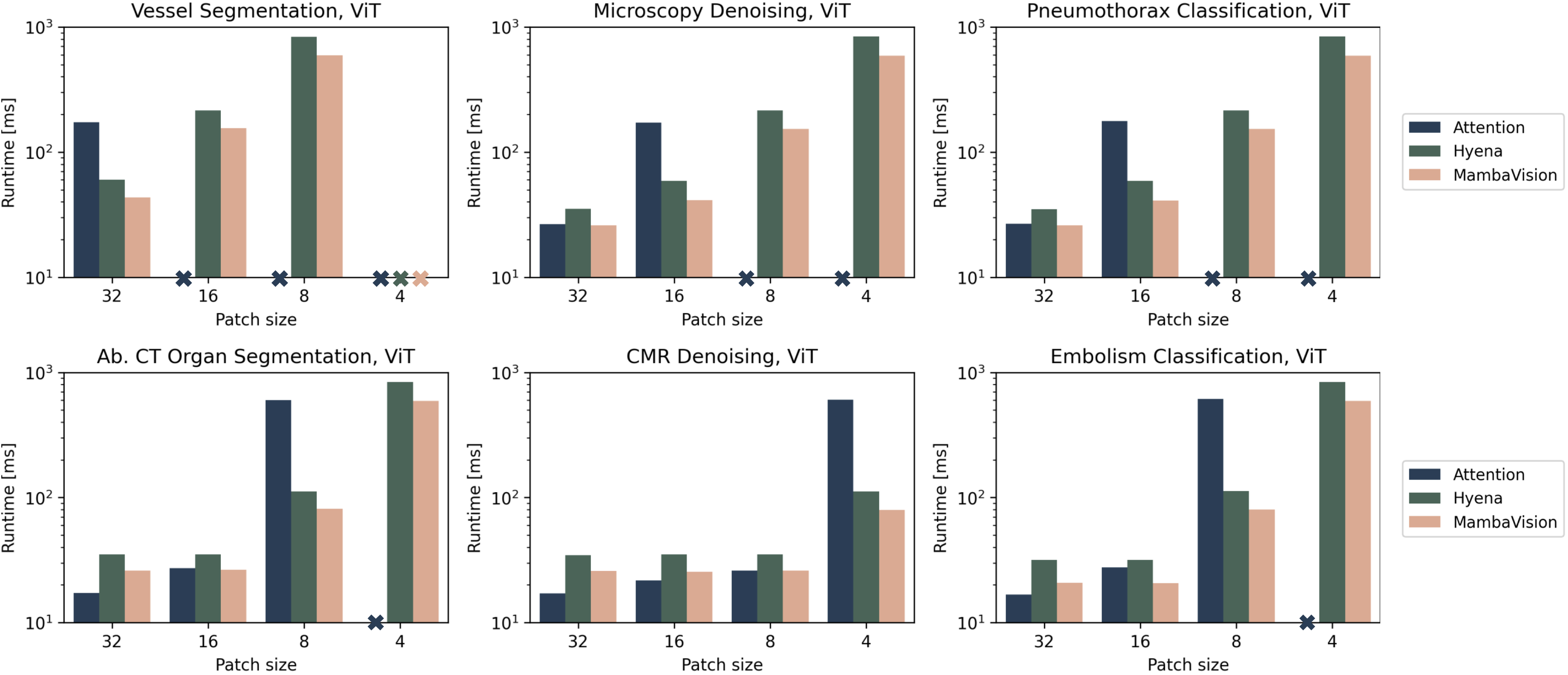}}
\end{figure*}

\begin{figure*}[htbp]
\floatconts
  {fig:swin_runtime}
  {\caption{Swin timing. We visualize timing for a forward and backward pass for each task, operator, and patch size. An X on the x-axis indicates that the window size exceeded our hardware capacity.}}
  {\includegraphics [width=.9\textwidth]{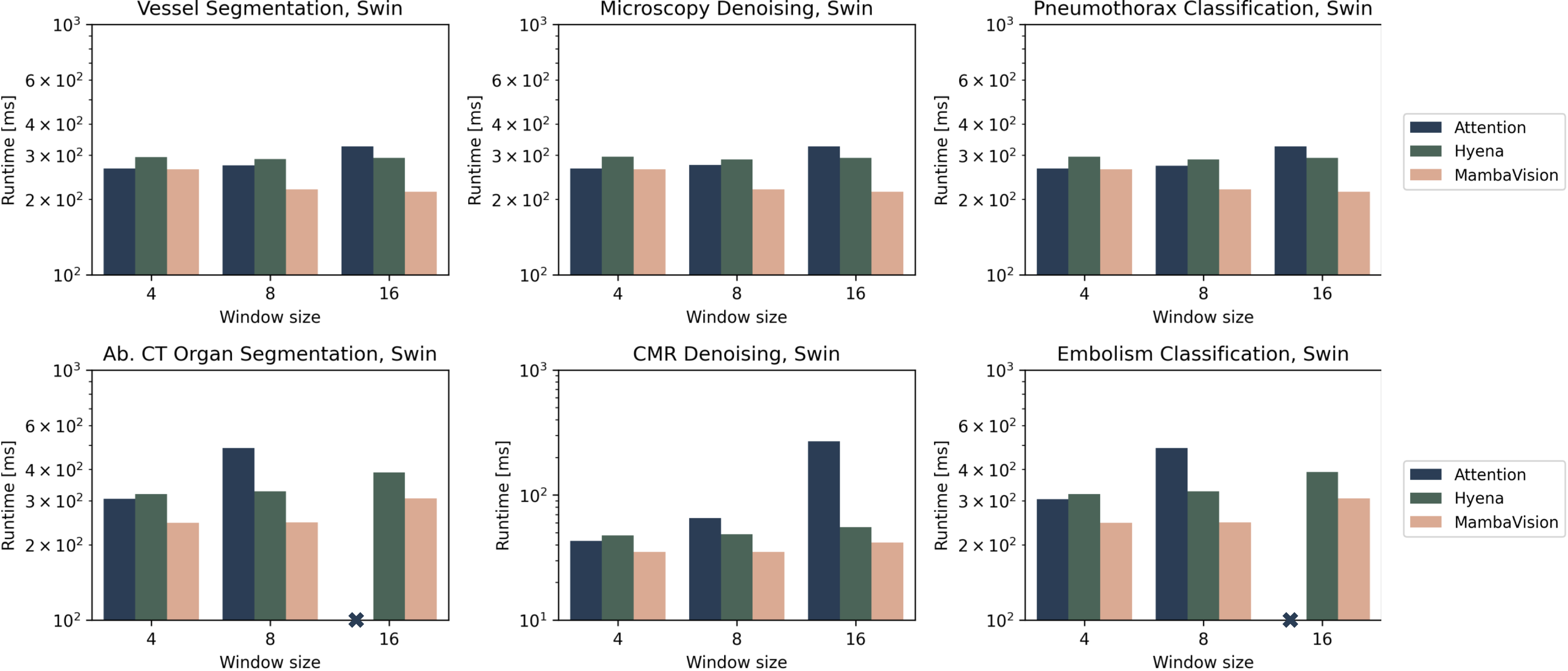}}
\end{figure*}

To assess runtime efficiency, we timed a forward and backward pass on a single NVIDIA A100 using a batch size of one. We only timed the backbone models (i.e., we did not include the linear, UNETR, or UPerNet task heads). We took the average of ten runs as the runtime reported in this work. We plot the runtime for each dataset and model configuration in Figures \ref{fig:vit_runtime} and \ref{fig:swin_runtime}. Note that the abdominal CT dataset and chest CT embolism dataset have approximately the same runtime and the chest x-ray pneumothorax dataset and the microscopy denoising dataset have approximately the same runtime due to these pairs of datasets having the same image sizes. For Swin, the vessels dataset also has the same runtime as the microscopy and chest x-ray datasets since it was resized to train the Swin models. 

\subsubsection{Maximum memory allocated}

\begin{figure*}[htbp]
\floatconts
  {fig:vit_memory}
  {\caption{ViT maximum memory allocated. We visualize maximum memory allocated for each task, operator, and patch size. An X on the x-axis indicates that the patch size exceeded our hardware capacity.}}
  {\includegraphics [width=.9\textwidth]{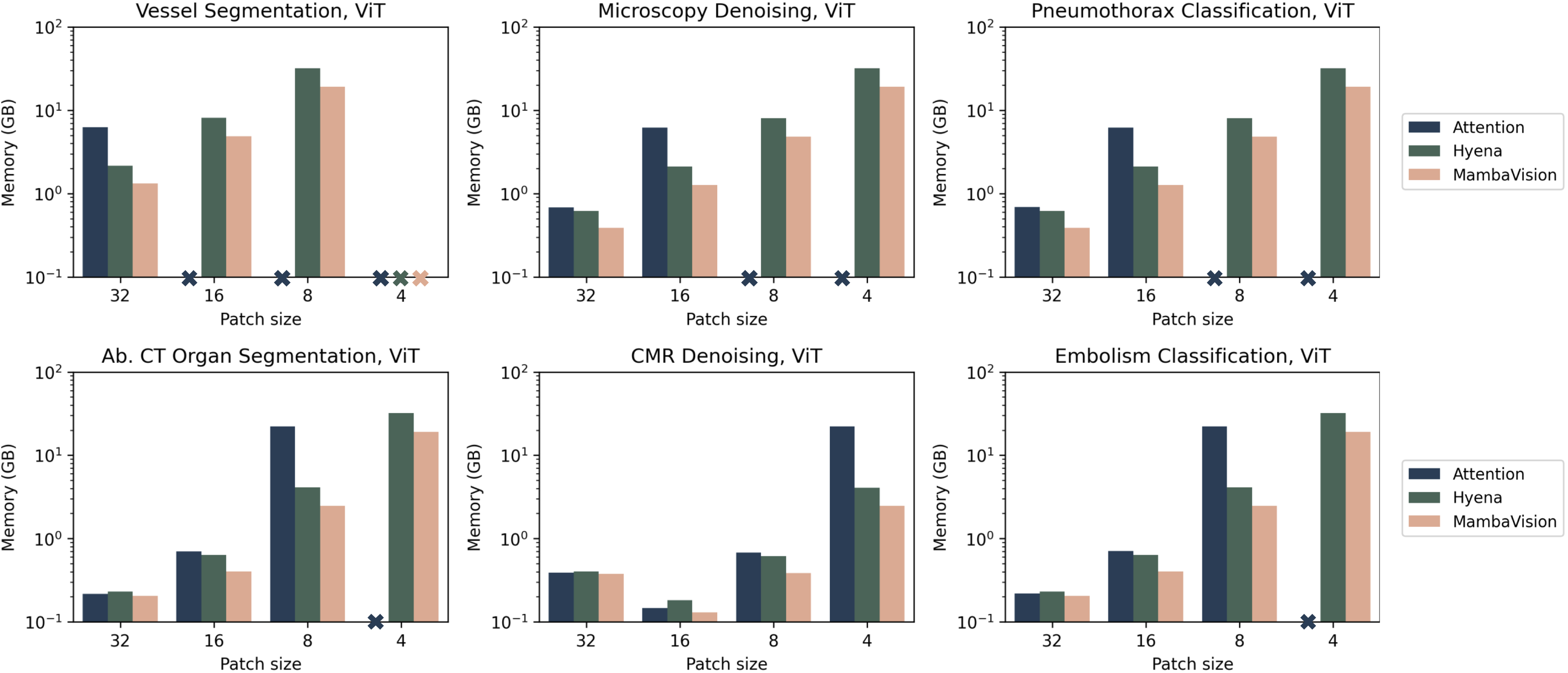}}
\end{figure*}

\begin{figure*}[htbp]
\floatconts
  {fig:swin_memory}
  {\caption{Swin maximum memory allocated. We visualize maximum memory allocated for each task, operator, and patch size. An X on the x-axis indicates that the window size exceeded our hardware capacity.}}
  {\includegraphics [width=.9\textwidth]{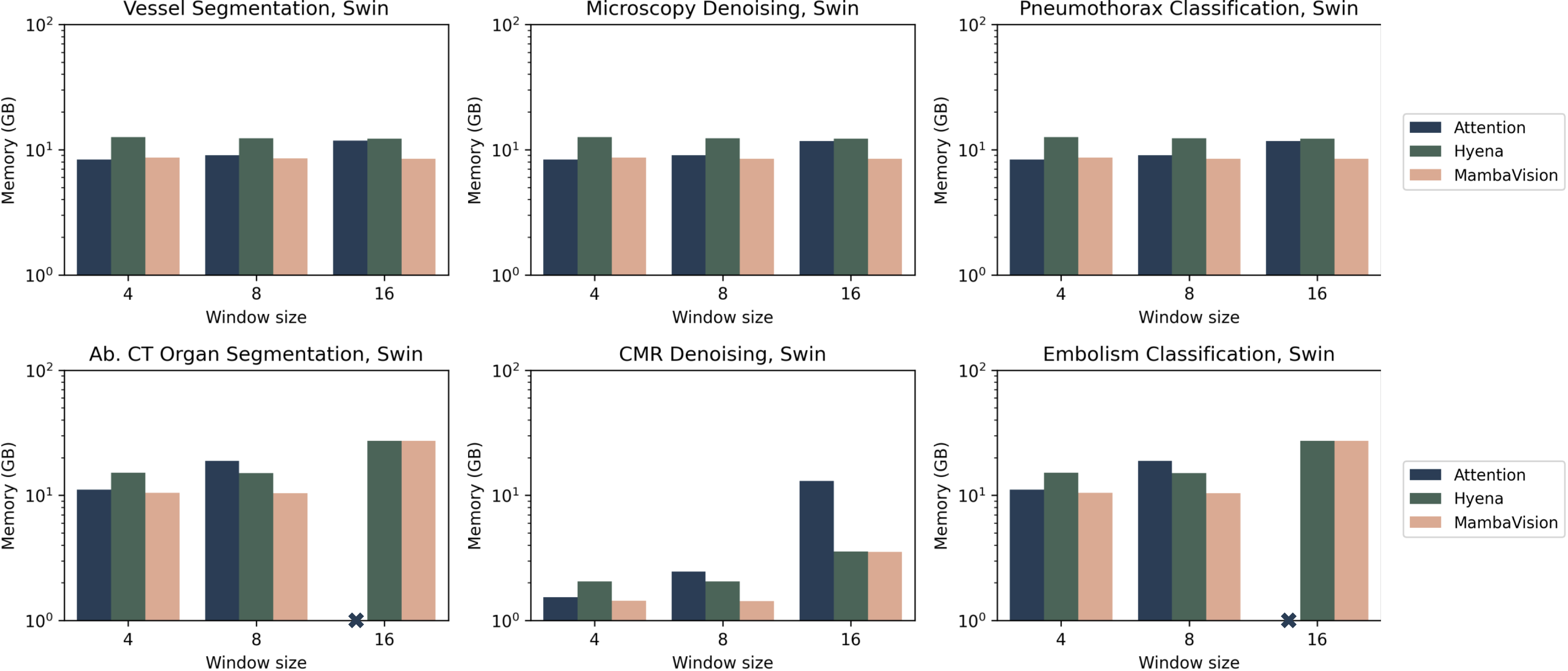}}
\end{figure*}

To assess memory efficiency, we recorded the maximum memory allocated on a single NVIDIA A100 using a batch size of one. We only assessed the backbone models (i.e., we did not include the linear, UNETR, or UPerNet task heads). We plot the maximum memory allocated for each dataset and model configuration in Figures \ref{fig:vit_memory} and \ref{fig:swin_memory}. Note that the abdominal CT dataset and chest CT embolism dataset have approximately the same memory and the chest x-ray pneumothorax dataset and the microscopy denoising dataset have approximately the same memory due to these pairs of datasets having the same image sizes. For Swin, the vessels dataset also has the same memory requirements as the microscopy and chest x-ray datasets since it was resized to train the Swin models. 

\subsection{Additional Results on Swin}

\subsubsection{Swin Patch Size}
In the main text, we discussed how context length can be varied by either changing the patch size or attention window. We varied patch size on ViT, while we kept the patch size constant for Swin and instead varied the attention window. In this section, we evaluate the impact of patch size on Swin performance.

Specifically, we investigated tokenizing the image with 4-pixel patches instead of 2-pixel patches (as used in the main text). We evaluated performance on all tasks using self-attention with a window size of eight and report the results in Table \ref{tab:swin_patch}. For segmentation, we report Dice; for denoising, we report SSIM; and for classification, we report AUROC. 95\% confidence intervals are reported in parentheses, computed by bootstrapping over the test set.

\begin{table}[hbtp]
\floatconts
  {tab:swin_patch}
  {\caption{Effect of patch size on Swin performance (95\% confidence intervals).}}
    {
    \centering
    \small
    \begin{tabular}{c|cc}
    \toprule
    & \shortstack{Patch\\4}
    & \shortstack{Patch\\2} \\
    \midrule
    Vessel & 
    0.85 (0.83-0.86) & 0.88 (0.87-0.89) \\
    Ab. CT & 
    0.80 (0.78-0.81) & 0.86 (0.84-0.87) \\
    Microscopy & 
    0.60 (0.55-0.64) & 0.60 (0.55-0.64) \\
    CMR & 
    0.50 (0.49-0.51) & 0.64 (0.64-0.65) \\
    Pneumothorax & 
    0.83 (0.81-0.85) & 0.86 (0.84-0.87) \\ 
    Embolism & 
    0.73 (0.70-0.76) & 0.79 (0.77-0.82) \\
    \bottomrule
    \end{tabular}}
\end{table}

We observe that smaller patches correspond to better performance. This is the same trend we observed in the main text with ViT, indicating that preserving resolution is important to achieving optimal performance in both architectures. 

\subsubsection{Window Shifting in Swin}

In the main text, we did not use window shifting when training the Swin transformers with Hyena or MambaVision. We opted not to use window shifting because doing so efficiently requires masking parts of the attention matrix; for additional details, see \cite{liu2021swin}. This masking operation does not have a straightforward analog for Hyena or MambaVision, so we removed the shift instead. We retained the shift operation when training the attention-based Swin networks to maintain the fidelity of the Swin transformer, as originally proposed.

To assess the impact of removing the shift operation, we report the results of training an attention-based Swin network with and without the shift operation. We trained these networks for all tasks and a window size of 8. We report results in Table \ref{tab:swin_shift}.

\begin{table}[hbtp]
\floatconts
  {tab:swin_shift}
  {\caption{Effect of window shifting on Swin performance (95\% confidence intervals).}}
    {
    \centering
    \small
    \begin{tabular}{c|cc}
    \toprule
    & \shortstack{Without \\ shift}
    & \shortstack{With \\ shift} \\
    \midrule
    Vessel & 
    0.88 (0.87-0.89) & 0.88 (0.87-0.89)  \\
    Ab. CT & 
    0.85 (0.84-0.87) & 0.86 (0.84-0.87) \\
    Microscopy & 
    0.60 (0.55-0.64) & 0.60 (0.55-0.64) \\
    CMR & 
    0.68 (0.67-0.68) & 0.64 (0.64-0.65) \\
    Pneumothorax & 
    0.78 (0.76-0.80) & 0.86 (0.84-0.87)  \\ 
    Embolism & 
    0.76 (0.73-0.79) & 0.79 (0.77-0.82) \\
    \bottomrule
    \end{tabular}}
\end{table}

We observe that only classification tasks experience degraded performance without the shift operation. In this case, an efficient implementation of Swin with shifting for the Hyena and MambaVision operators may further boost their performance on classification tasks. We note that this shift operation may explain the performance difference between Swin classification using self-attention vs. the alternative operators in the main text.

\end{document}